\documentclass[10pt]{article}
\PassOptionsToPackage{hyphens}{url}   

\usepackage{amsmath}
\usepackage{scicom}
\usepackage[authoryear]{natbib}

\usepackage{hyperref}
\usepackage{bookmark}
\usepackage{url}
\usepackage{nicefrac}
\usepackage{microtype}
\usepackage{tabularx}
\usepackage{multirow}
\usepackage{float}
\usepackage{adjustbox}
\usepackage{longtable}
\usepackage{array}
\usepackage[section]{placeins}

\newcolumntype{L}[1]{>{\raggedright\arraybackslash}p{#1}}

\hypersetup{
pdftitle={How to Reduce Whisper Hallucination},
pdfauthor={Husein Zolkepli},
pdfsubject={Audio and Speech Processing},
pdfkeywords={speech recognition, hallucination, whisper, fine-tuning},
colorlinks=true,
linkcolor=scLink,
citecolor=scLink,
urlcolor=scLink
}
\title{How to Reduce Whisper Hallucination}
\scshorttitle{How to Reduce Whisper Hallucination}
\scauthor{Husein Zolkepli}{husein.zolkepli@scicom.com.my}
\scaffiliation{Scicom (MSC) Berhad, Malaysia}
\scdate{October 2026}

\begin{document}

\maketitle

\begin{abstract}
Whisper changed what open speech recognition means: one permissively licensed checkpoint, 99
languages, no per-language tuning, running on a laptop. Four years on it is still what runs in
production, downloaded roughly 4.6 M times a month against 568 k for its nearest open rival,
and it is still the teacher that Distil-Whisper, Kotoba-Whisper and NVIDIA's Granary are
distilled from. But it writes sentences nobody said. On 42 clips of pure room tone, \texttt{whisper-large-v3}
emits words on 61.9\% of them and emits something on 100\% of them. The usual response is to
distil a student from Whisper pseudo-labels and filter the hallucinations out of the corpus
first. That deletes the evidence while leaving the behaviour in place: the student is fitted to the
subset where the teacher was right, and inherits a failure mode that never appears in its own
training data. The second response, adding non-speech audio so the model learns to stay quiet, is
already used in production systems and it works, but it teaches suppression without teaching
discrimination. The checkpoint that is best on every non-speech arm here is also the one that
recovers the fewest genuinely spoken phrases and deletes 58\% of repeated speech. We argue the teacher has to be fixed, with both halves of the signal, and we do the work that
requires. Across 33 matched pairs of fine-tunes that differ only by those synthetic positives, adding
them lowers word emission on real voice-free audio in 31 pairs and raises phrase recovery in
32. Among the 30 pairs that hold English accuracy, it is 30 out of 30. The best checkpoint
takes hallucination on silence from 61.9\% to 2.4\% and words over real voice-free audio from
99.9\% to 47.8\% while raising phrase recovery from 69.8\% to 82.7\%. We build a
benchmark that scores both halves of the problem at once, 11,852 clips over eight arms plus
6,267 synthetic positives, 8,296 clips of real audio that made a production model fail, and
FLEURS in 58 languages. We collect a lexicon of 40,891 hallucination phrases in 100 languages,
select a text-to-speech system by measurement rather than reputation, and synthesise those
phrases as positives so the model sees the same text both as something to suppress and as
something to transcribe. We then sweep fine-tuning over rank, learning rate, and the presence
of the synthetic positives. The benchmark, the lexicon and the synthetic corpus are released at
\href{https://huggingface.co/datasets/Scicom-intl/Whisper-Hallucination}{Scicom-intl/Whisper-Hallucination}.
\end{abstract}

\section{Introduction}

Whisper \citep{radford2022whisper} changed what open speech recognition means. Multilingual
transcription used to mean a model per language and a pipeline per domain. Whisper made it one permissively licensed
checkpoint covering 99 languages, with no per-language tuning, running on a laptop. Four years later it is still the default, and it is good: we measured it on 20 FLEURS
\citep{conneau2022fleurs} test clips in each of 58 languages, and
\texttt{whisper-large-v3} reaches a character error rate of 0.075 on the typical language,
with 35 of the 58 under 0.15.

This paper is about the one thing it does badly, and it is worth being clear at the outset
that the thing it does badly is narrow. Whisper is not broken. It transcribes speech extremely
well. It just does not know how to say nothing.

More than that, Whisper is the teacher. Distil-Whisper \citep{gandhi2023distilwhisper} is
trained on Whisper pseudo-labels. So is Kotoba-Whisper \citep{kotobawhisper2024}. So is
Granary \citep{koluguri2025granary}, and therefore so are Canary-1B-v2 and
Parakeet-TDT-0.6B-v3 \citep{sekoyan2025canaryv2}. A defect in Whisper is not contained to
Whisper. It propagates into the corpora that train its successors.

It is no longer the most accurate model, and it is still the one in production. Canary-1B-v2
reports an average word error rate of 8.1\% against \texttt{whisper-large-v3}'s 9.9\% on its
25 European languages. SeamlessM4T \citep{seamless2023m4t} reported a 45\% relative reduction
over \texttt{whisper-large-v2} on FLEURS across 77 overlapping languages. Omnilingual ASR
\citep{keren2025omnilingual} covers more than 1,600 languages. None of that has displaced it,
for four reasons that have nothing to do with word error rate.

\textbf{Coverage.} This is the big one. Canary-1B-v2 and Parakeet-TDT-0.6B-v3 cover 25
languages, all but two of them European. Whisper covers 99. For anyone working outside that
list, an 8.1\% average on 25 languages is not an improvement on 9.9\%, it is a model that
does not run. The 1.8-point gap only exists where both models work, and most of the world's
speakers are not in that intersection.

\textbf{Licence.} Whisper's code is MIT and its weights are Apache-2.0, so it ships in a
commercial product without a conversation. SeamlessM4T, the model with the 45\% reduction, is
CC-BY-NC: that number is not available to anyone building a product. Canary-1B-v2 and
Parakeet-TDT-0.6B-v3 are CC-BY-4.0 and are genuinely usable, within those 25 languages.

\textbf{Code-switching.} Real speech in most of the world mixes languages inside one
utterance. The corpus behind this paper is Malaysian, where a single sentence routinely moves
between Malay, English, Mandarin and Tamil, and Whisper is the model practitioners reach for
because it degrades gracefully when the language changes mid-clip rather than locking to the
language it detected first.

\textbf{Adoption.} At the time of writing \texttt{whisper-large-v3} is downloaded roughly
4.6 M times a month from Hugging Face, against 568 k for Parakeet-TDT-0.6B-v3, 301 k for
SeamlessM4T v2 and 55 k for Canary-1B-v2. Omnilingual ASR is recent enough that the only
copies on the Hub are community conversions. Whatever the leaderboards say, Whisper is what is
running.

So the defect sits in the model most production systems use, and in the teacher that
Distil-Whisper, Kotoba-Whisper and Granary distilled from. Section~\ref{sec:distil} argues that the way those corpora are built
lets the behaviour through while deleting the evidence for it.

The failure is this. Over audio with no speech in it, Whisper writes a sentence anyway. Not
rarely, not only on hard audio, and not harmlessly.
\citet{koenecke2024careless} ran more than 13,000 clips from AphasiaBank through Whisper and
found that roughly 1\% of transcriptions contained entire hallucinated phrases or sentences
present nowhere in the audio, and that 38\% of those hallucinations carried explicit harms:
perpetuating violence, inventing associations, or implying false authority. Hallucination rates
were higher for speakers with longer non-vocal pauses, which is a symptom of aphasia, so the
failure is not distributed evenly across speakers.
\citet{baranski2025nonspeech} come at the same problem from the audio side and show that a
curated list of common hallucinations works as a post-processing filter. A recogniser that
writes a plausible sentence over silence is worse than one that returns an error, because
nothing downstream can tell the difference.

The five checkpoints measured throughout this paper are named as follows. Three are the
OpenAI releases. The other two are Mesolitica's Malay fine-tunes, which matter here because
their training pipeline is public and already contains non-speech audio, so they are the
closest existing thing to the fix this paper proposes. Table~\ref{tab:models} names all five
and fixes the short labels used in every table and figure below.

\begin{table}[!htbp]
\centering
\caption{The checkpoints, and the short labels used in every table and figure below. The two
Malay fine-tunes are built on \texttt{large-v2} and \texttt{turbo} respectively, and
\texttt{turbo} differs from \texttt{large-v3} in having four decoder layers rather than
thirty-two.}
\label{tab:models}
\footnotesize
\begin{tabular}{lllL{0.16\linewidth}}
\toprule
\schead \scth{label} & \multicolumn{2}{l}{\scth{identifier}} & \scth{what it is} \\
\midrule
\texttt{large-v2} & \texttt{openai/} & \texttt{whisper-large-v2} & base, 1.55 B \\
\texttt{large-v3} & \texttt{openai/} & \texttt{whisper-large-v3} & base, 1.55 B \\
\texttt{turbo} & \texttt{openai/} & \texttt{whisper-large-v3-turbo} & base, 4 layers \\
\texttt{malaysian-v2} & \texttt{mesolitica/} & \texttt{malaysian-whisper-large-v2} & Malay tune \\
\texttt{malaysian-turbo-v3} & \texttt{mesolitica/} & \texttt{Malaysian-whisper-large-v3-turbo-v3} & Malay tune \\
\bottomrule
\end{tabular}
\end{table}

Table~\ref{tab:nonspeech} measures it on 1,810 clips that contain no speech at all: 42 of
room tone, 600 of music, 1,168 of label-verified non-speech sound.

\begin{table}[!htbp]
\centering
\caption{Hallucination on audio that contains no speech. The correct output is the empty
string, so any output at all is wrong. The first block counts outputs containing words, the
second counts any output including a bare full stop.}
\label{tab:nonspeech}
\footnotesize
\begin{tabular}{lrrrrrr}
\toprule
\schead \scth{arm} & \scth{n} & \scth{large-v2} & \scth{large-v3} & \scth{turbo}
 & \shortstack{\scth{malaysian}\\\scth{-v2}}
 & \shortstack{\scth{malaysian}\\\scth{-turbo-v3}} \\
\midrule
\texttt{silence} & 42 & 85.7\% & 61.9\% & 59.5\% & 35.7\% & \textbf{16.7\%} \\
\texttt{music} & 600 & 98.7\% & 97.0\% & 96.7\% & 69.7\% & \textbf{31.0\%} \\
\texttt{nonspeech} & 1{,}168 & 98.8\% & 89.9\% & 80.7\% & 76.0\% & \textbf{9.3\%} \\
\midrule
\texttt{silence} any output & 42 & 100\% & 100\% & 100\% & 64.3\% & \textbf{16.7\%} \\
\texttt{music} any output & 600 & 100\% & 100\% & 100\% & 73.0\% & \textbf{31.7\%} \\
\texttt{nonspeech} any output & 1{,}168 & 100\% & 100\% & 100\% & 77.8\% & \textbf{9.4\%} \\
\bottomrule
\end{tabular}
\end{table}

Every OpenAI checkpoint in Table~\ref{tab:nonspeech} writes something on every single clip
that has no speech in it. Read
the first block and the second block together, because published hallucination rates for the
same model on the same audio differ by more than ten times depending on which one an author
chose to report.

The second failure is repetition. On 1,440 clips built from a known unit repeated a known
number of times, the 95th percentile of emitted repeats divided by true repeats is 1.00 for
\texttt{large-v3} and 18.3 for \texttt{turbo}. The two Malay fine-tunes reach 55.0 and 73.3.
A single clip can produce hundreds of copies of one token.

Both failures come from the same place. The Whisper decoder is a language model conditioned on
an audio encoder. When the acoustic evidence is weak or absent, the prior takes over and the
decoder writes fluent text, because writing fluent text is what it was trained to do. Silence
is not a token it can emit. It has to decide to stop, and stopping is exactly what it is worst
at.

The field's response has been to work around the model. Put a voice activity detector in front
of it, threshold on \texttt{no\_speech\_prob}, block a list of known phrases after the fact.
All of these help, and none of them change the weights. Move to a serving stack that does not
implement Whisper's own decoding loop, which is what happens in production, and most of the
knobs disappear. vLLM exposes no \texttt{no\_speech\_threshold}, no
\texttt{compression\_ratio\_threshold}, no \texttt{logprob\_threshold}, and no
\texttt{condition\_on\_previous\_text}. What is left is the model.

\section{Distilling from Whisper}
\label{sec:distil}

The standard way to get a cheaper or a more multilingual Whisper is to run Whisper over
unlabelled audio, keep the transcripts, and train a student on them. The pipelines are public
and so are the filters, which is what makes this arguable rather than speculative.

Distil-Whisper \citep{gandhi2023distilwhisper} pseudo-labels with Whisper and then, in its own
words, uses ``a simple word error rate heuristic'' to select only the highest quality
pseudo-labels for training. Kotoba-Whisper \citep{kotobawhisper2024} applies the same recipe to
Japanese with \texttt{large-v3} as the teacher. NVIDIA's Granary
\citep{koluguri2025granary} builds a 25 language corpus with a pseudo-labeling pipeline whose
stages are listed as ``segmentation, two-pass inference, hallucination filtering, and
punctuation restoration'', and Canary-1B-v2 and Parakeet-TDT-0.6B-v3
\citep{sekoyan2025canaryv2} are trained on it.

Read those two descriptions next to each other. A hallucination filter is a device for removing
the teacher's hallucinations from the student's training set. It works. The corpus comes out
cleaner, and the resulting models are strong. But the examples it deletes are precisely the
ones that demonstrate the failure, so the student is fitted to the subset of the teacher's
behaviour that was already correct. It never sees a clip where the teacher wrote
\texttt{thank you for watching} over a music bed. It has no way to learn that this is wrong. It
learns the teacher's mapping on clean speech and inherits the teacher's prior everywhere else.

There is a second, quieter effect. Pseudo-label pipelines are built from speech corpora, and
speech corpora are segmented to speech. A clip with no voice in it yields a pseudo-label that
looks like garbage, so the filter drops it, or it was never in the corpus in the first place.
The student therefore has close to zero training examples whose correct output is the empty
string. Asking it to say nothing at inference time is asking for behaviour with no support in
its training distribution.

We measured how strongly segmentation hides this. GigaSpeech \citep{chen2021gigaspeech} is
podcasts and YouTube, which is the right material, and mining it for hallucinations yields
0.41\% of clips. AudioSet \citep{gemmeke2017audioset}, the same kind of source but unsegmented,
yields 54.70\%. The failures live in what segmentation throws away.

\subsection{Negative-only training}

The obvious remedy is to put non-speech audio back in. Two groups already do.

Canary-1B-v2 \citep{sekoyan2025canaryv2} states it plainly: the training mix has ``non-speech
audio added to reduce hallucinations for ASR and AST''. Mesolitica's Malaysian Whisper
fine-tunes \citep{mesolitica2025malaysianwhisper} do the same thing and, unusually, publish the
whole pipeline. Their stage-two training set \citep{mesolitica2025stage2} lists an AudioSet
block of 805k rows with a dedicated \texttt{noise} subset of 13.4k, and the preparation code
is open \citep{malaysiaai2025pipeline}.

It works, and our benchmark shows exactly how well. In Table~\ref{tab:nonspeech},
\texttt{Malaysian-turbo-v3} is the best checkpoint measured on every non-speech arm, by a wide
margin: 16.7\% on silence against 61.9\% for base \texttt{large-v3}, and 9.3\% on non-speech
against 89.9\%. Adding negatives is the single most effective thing anyone in this table did.

Then look at what it cost. The same checkpoint recovers 41.1\% of phrases that are genuinely
spoken, against 69.8\% for the base model it was tuned from, deletes
58.0\% of clips containing repeated speech, and returns character error rate 1.26 on the typical
FLEURS language. It did not learn to detect silence. It learned a prior against emitting text,
and that prior fires on real speech too.

This is the gap we set out to close. Negative-only supervision teaches suppression without
teaching discrimination. The missing ingredient is audio that genuinely contains the phrases the
negatives are teaching the model to suppress: a recording where somebody really does say
\texttt{thank you for watching}, labelled with that transcript. No public pipeline has one,
because nobody collected the phrases first. So the fix has to be applied to the teacher, and it
has to supply both halves.

\section{Requirements}

Three things, in order.

\textbf{A benchmark that scores both halves.} A model that emits nothing is perfect on every
non-speech arm and useless. Any metric that looks only at hallucination will select it. The
benchmark has to measure suppression and accuracy on the same axis, on the same run.

\textbf{Knowledge of what the model actually says.} Hallucinations are not random. They are a
small set of phrases the model falls back on, and they differ by language. That set is
enumerable, and once enumerated it can be turned into training data.

\textbf{Training data where both answers appear.} The model needs clips whose correct output is
the empty string, and it needs clips where the very same phrases are genuinely spoken and must
be transcribed. Without the second half, suppression is learned as a blanket prior against
emitting text, which is the failure mode of every aggressive fine-tune we measured.

\section{Dataset}

\subsection{Non-speech audio}

Table~\ref{tab:arms} lists the eight arms, what each is built from and what the correct output
is for it. Three of them have the empty string as the answer, which is what makes them able to
measure invention at all.

\begin{table}[!htbp]
\centering
\caption{The eight benchmark arms. Every arm is a \texttt{test} split. The training corpus is a
separate build that is disjoint by construction, verified before any training run.}
\label{tab:arms}
\footnotesize
\begin{tabular}{lL{0.34\linewidth}rl}
\toprule
\schead \scth{arm} & \scth{source} & \scth{clips} & \scth{correct output} \\
\midrule
\texttt{silence} & 6 noise floors $\times$ 7 durations & 42 & nothing \\
\texttt{music} & Free Music Archive \citep{defferrard2017fma} & 600 & nothing \\
\texttt{nonspeech} & FSD50K \citep{fonseca2020fsd50k}, label-verified voice-free & 1{,}168 & nothing \\
\texttt{reduplication} & a unit repeated an exact number of times & 1{,}440 & that many repeats \\
\texttt{speech\_in\_noise} & genuine speech mixed at 5 SNRs & 1{,}200 & the transcript \\
\texttt{genuine} & real Malaysian speech & 4{,}694 & the transcript \\
\texttt{genuine\_isolated} & one phrase spoken alone & 88 & the phrase \\
\texttt{librispeech\_test\_clean} & LibriSpeech \citep{panayotov2015librispeech} & 2{,}620 & the transcript \\
\bottomrule
\end{tabular}
\end{table}

The \texttt{reduplication} arm deserves a note, because it is the only arm where both extremes
are wrong. A clip contains a unit repeated exactly $n$ times. Emitting more is runaway. Emitting
nothing is deletion. A metric that only penalises runaway will reward a model that deletes
repeated speech, and one of our own fine-tunes does exactly that.

\subsection{Spoken positives}

The negative arms ask what a model invents over silence. The positive arm asks what it does
when the same phrase is real. This is the half that stops the benchmark from selecting a mute
model, and it is built by synthesising the lexicon described in Section~\ref{sec:lexicon}.
It contains 29,112 clips in 83 languages, split into 22,845 train and 6,267 test.

Each test clip is scored three ways, because production audio arrives as a chunk with a noise
floor around it, not as a bare 0.9 second clip. The conditions are the clip as published, the
clip with two seconds of digital zeros on each side, and the clip with two seconds of real room
tone on each side drawn from the \texttt{silence} arm. The zeros condition is a control and
should do nothing, because Whisper pads every input to a 30 second window anyway.

\subsection{Wild audio}

Everything above is a built stimulus. The wild pool is real recordings that made a model fail:
3,607 Earnings-22 \citep{delrio2022earnings22} clips carrying human span annotations, 4,689
clips mined from streamed corpora with no annotator, and 187 aphasia clips kept local because
the source is membership-gated. It ships as 3,628 train and 4,668 test, split by source
recording rather than by clip, because a dozen segments can come from one call and a clip-level
split would leak near neighbours.

Mining needs no annotator. Two signatures are self-evident: a token run of six or more, and
words emitted where Silero voice activity detection finds no speech. We never call a clip a
hallucination because a transcript disagrees with a reference, since that finds ordinary
recognition errors. Table~\ref{tab:yield} gives the yield corpus by corpus, and the spread
across it is the reason the pool is built the way it is.

\begin{table}[!htbp]
\centering
\caption{Hallucination yield by corpus, per 8,000 clips heard. Recording quality predicts yield
across five thousandfold.}
\label{tab:yield}
\footnotesize
\begin{tabular}{llrr}
\toprule
\schead \scth{corpus} & \scth{character} & \scth{kept} & \scth{rate} \\
\midrule
AudioSet \citep{gemmeke2017audioset} & YouTube, heavy background noise & \textbf{4{,}376} & \textbf{54.70\%} \\
AMI \citep{carletta2005ami} & spontaneous meetings, far-field & 235 & 2.94\% \\
GigaSpeech \citep{chen2021gigaspeech} & podcasts and YouTube, speech-aligned & 33 & 0.41\% \\
Earnings-22 \citep{delrio2022earnings22} & conference calls & 31 & 0.39\% \\
People's Speech \texttt{dirty} \citep{galvez2021peoplesspeech} & noisy transcripts, clean audio & 9 & 0.11\% \\
VoxPopuli \citep{wang2021voxpopuli} & parliament & 4 & 0.05\% \\
People's Speech \texttt{clean} & curated read speech & 1 & 0.01\% \\
\bottomrule
\end{tabular}
\end{table}

Two lessons. Curated corpora barely fail, so a hallucination
study built on them will conclude the problem is small. And a corpus can be the right material and still hide the failure:
GigaSpeech is podcasts and YouTube and yields 0.41\%, because its segments are cut to speech.

\subsection{Multilingual accuracy}

\texttt{librispeech\_test\_clean} is English. A multilingual fine-tune can hold its English word
error rate and destroy everything else, and nothing in an English benchmark will show it. We
add a FLEURS arm: 20 test clips in each of 58 languages, averaged per language rather than per
clip so a large language cannot mask a collapsed one, and in character error rate because
Chinese, Japanese, Thai, Lao, Burmese and Khmer do not put spaces between words.

This arm paid for itself immediately, as Section~\ref{sec:bench} shows.

\section{Benchmark results}
\label{sec:bench}

Five checkpoints, greedy decoding, no forced language, no temperature fallback. Every
checkpoint is run on all of it, roughly 40,000 clips each.

Three are the OpenAI base models. The other two are Mesolitica's Malaysian fine-tunes, and they
are in the set for a reason: they are the closest public thing to the fix this paper proposes.
Their pipeline is open, their training mix already contains non-speech audio, and they are the
only checkpoints here that were deliberately tuned against hallucination. Treat them as the
strong baseline, not as a foil.

\begin{figure}[H]
  \centering
  \includegraphics[width=0.78\linewidth]{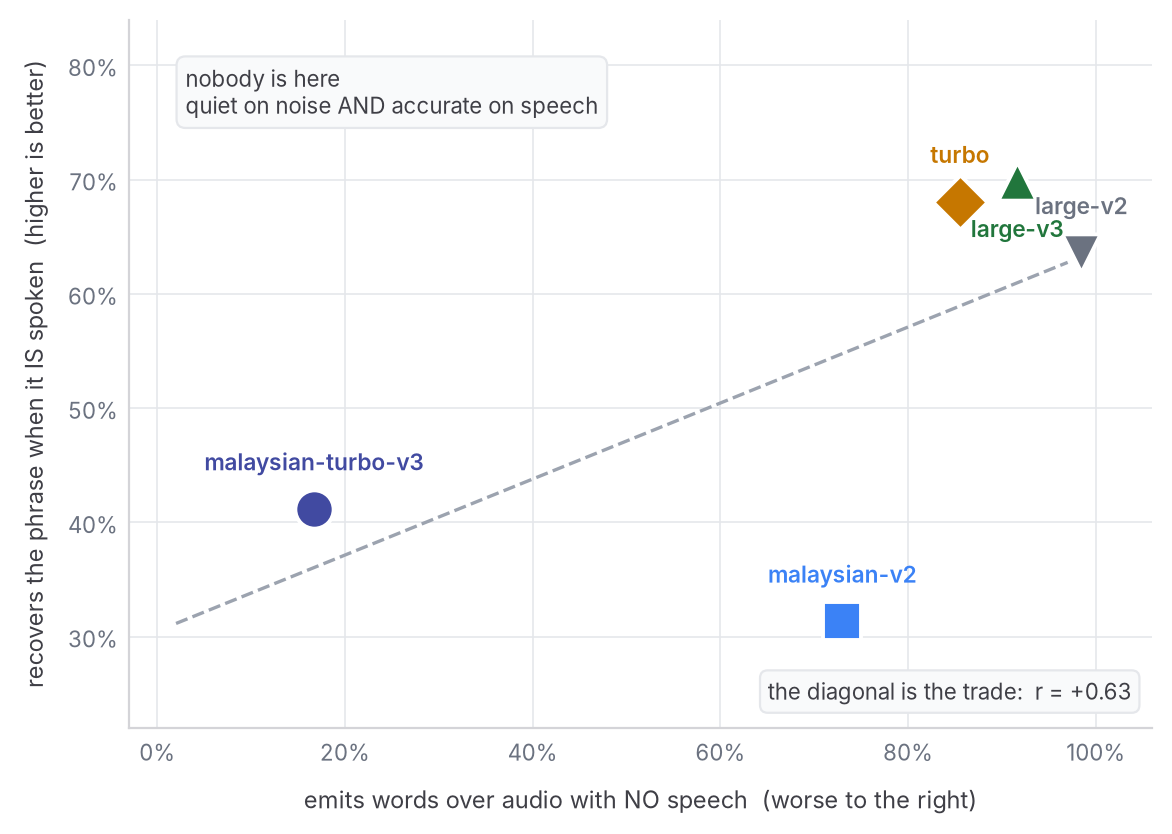}
  \caption{Every checkpoint trades one failure for the other. Horizontal axis is how often the model emits words over audio with no speech in it, pooled over the three non-speech arms and weighted by clip count. Vertical axis is how often it recovers a phrase that is genuinely spoken. The top left corner is the model everyone wants and nothing is there.}
  \label{fig:tradeoff}
\end{figure}

Figure~\ref{fig:tradeoff} puts the two halves on one pair of axes, invention on the
horizontal and recovery on the vertical, so a checkpoint's position is its whole character.

\begin{figure}[H]
  \centering
  \includegraphics[width=0.78\linewidth]{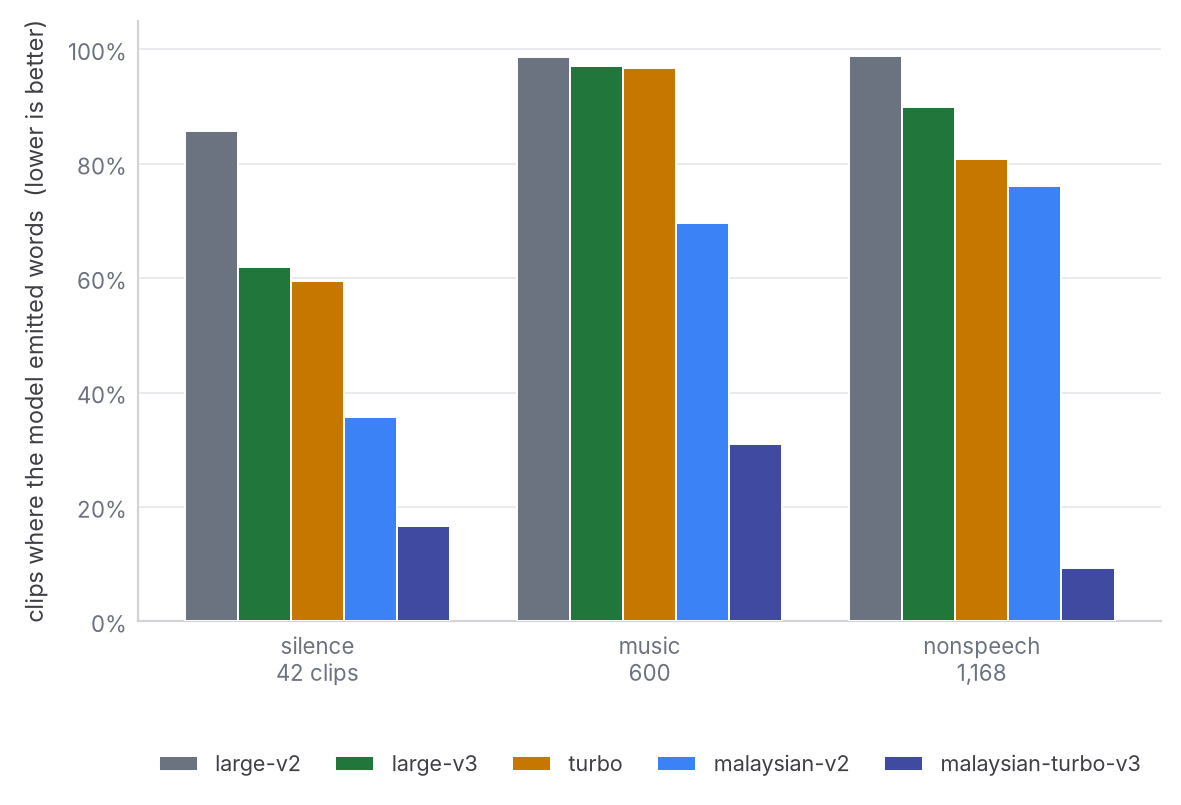}
  \caption{Hallucination on the three non-speech arms. Every bar is error, because the correct output is the empty string.}
  \label{fig:benchmark}
\end{figure}

\textbf{No checkpoint is both quiet on noise and accurate on speech.} The five sit on a line,
$r = +0.63$ between the pooled non-speech word-emission rate and phrase recovery. Quiet models delete real speech and accurate models write over silence. This is
the central measurement of the paper and it is what makes a one-sided metric dangerous.
Figure~\ref{fig:benchmark} breaks the non-speech half out arm by arm, and the ordering of the
five checkpoints holds across silence, music and non-speech sound alike.

\begin{figure}[H]
  \centering
  \includegraphics[width=0.78\linewidth]{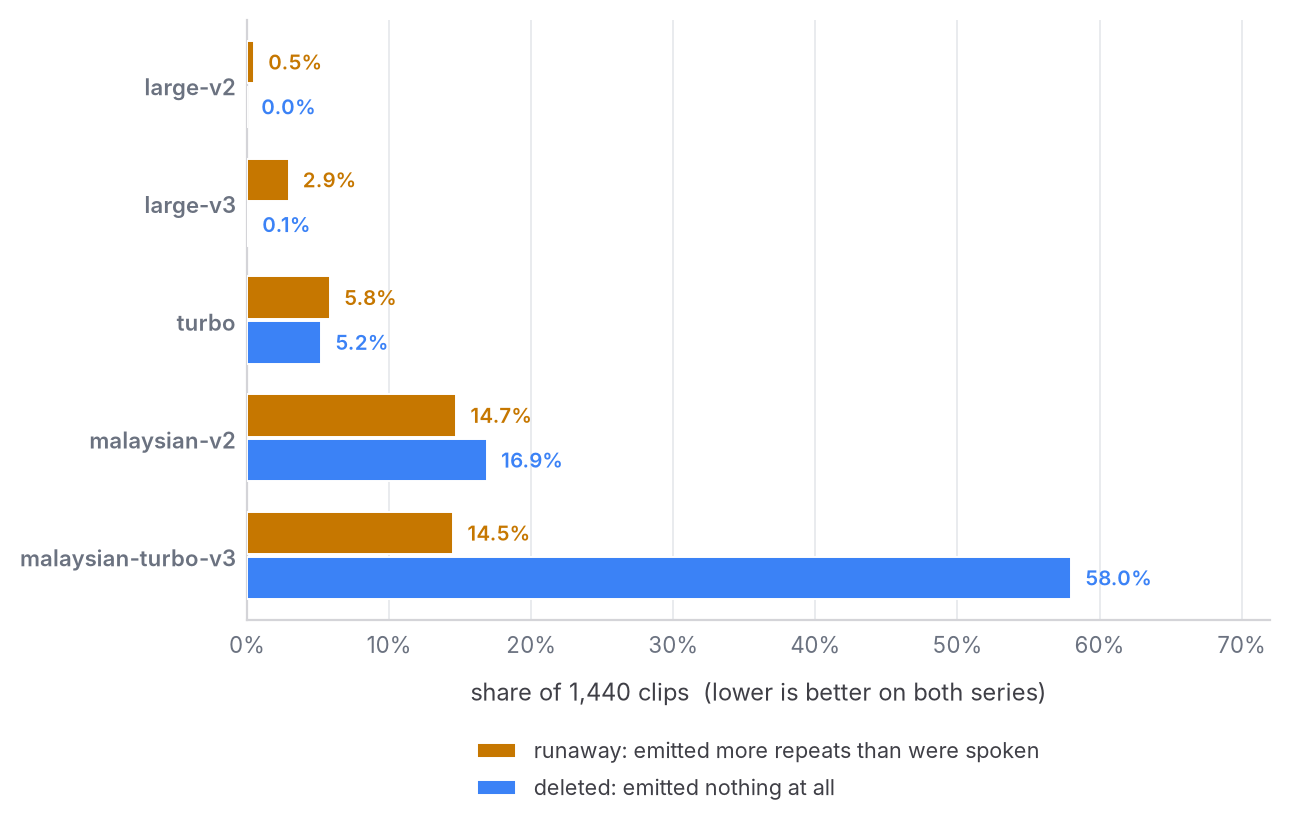}
  \caption{Both extremes are wrong on repeated speech. A model that emits more repeats than were spoken is running away; a model that emits nothing has deleted the speech. \texttt{Malaysian-turbo-v3} does the second on 58.0\% of clips.}
  \label{fig:redup}
\end{figure}

Figure~\ref{fig:redup} shows both failure modes on the reduplication arm at once, running
away and deleting, because a model can only be judged on this arm by looking at both.

Table~\ref{tab:wer} is the other half of the ledger, where a model that says nothing has
nowhere to hide.

\begin{table}[!htbp]
\centering
\caption{Accuracy. The last two rows are the multilingual guard, in character error rate.}
\label{tab:wer}
\footnotesize
\begin{tabular}{lrrrrrr}
\toprule
\schead \scth{arm} & \scth{n} & \scth{large-v2} & \scth{large-v3} & \scth{turbo}
 & \shortstack{\scth{malaysian}\\\scth{-v2}}
 & \shortstack{\scth{malaysian}\\\scth{-turbo-v3}} \\
\midrule
\texttt{librispeech} WER & 2{,}620 & 4.5\% & \textbf{3.5\%} & \textbf{3.5\%} & \textbf{3.5\%} & 10.6\% \\
\texttt{genuine} WER & 4{,}694 & 35.4\% & \textbf{31.8\%} & \textbf{31.8\%} & 42.3\% & 60.7\% \\
\texttt{speech\_in\_noise} WER & 1{,}200 & 41.6\% & \textbf{34.5\%} & 36.1\% & 51.1\% & 58.7\% \\
\midrule
\texttt{fleurs} CER, typical language & 58 langs & 0.097 & \textbf{0.075} & 0.078 & 1.051 & 1.259 \\
\texttt{fleurs} CER, macro mean & 1{,}160 & 0.422 & \textbf{0.305} & 0.511 & 1.781 & 1.681 \\
\bottomrule
\end{tabular}
\end{table}

\begin{figure}[H]
  \centering
  \includegraphics[width=0.78\linewidth]{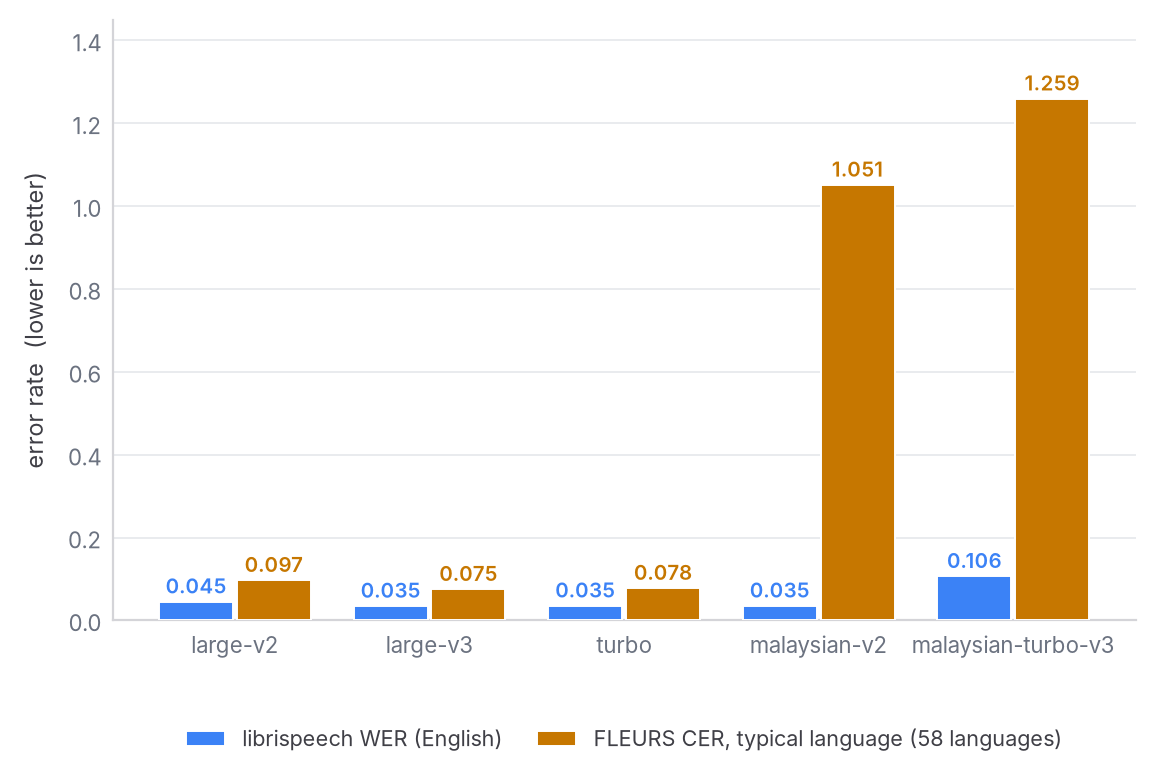}
  \caption{English accuracy hides what a multilingual fine-tune destroyed. \texttt{malaysian-whisper-v2} matches base \texttt{large-v3} on LibriSpeech and returns character error rate 1.05 on the typical FLEURS language.}
  \label{fig:accuracy}
\end{figure}

\textbf{English accuracy says nothing about the other 57 languages.}
\texttt{malaysian-whisper-v2} matches base \texttt{large-v3} on LibriSpeech, 3.5\% either way,
and returns character error rate 1.05 on the typical FLEURS language. Two of its 58 languages
land under 0.15, against 35 for \texttt{large-v3}. Specialising on Malay cost everything else,
and no English benchmark can see it (Figure~\ref{fig:accuracy}).

The macro mean and the median disagree by four times because Whisper cannot transcribe some of
these languages at all. For base \texttt{large-v3}, Amharic sits at 1.81 character error rate,
Khmer at 1.58, Somali at 1.44, while Spanish, Italian and Dutch sit at 0.01. The mean measures
the tail and the median measures the typical language. Quoting one without the other misleads.

\begin{figure}[H]
  \centering
  \includegraphics[width=0.78\linewidth]{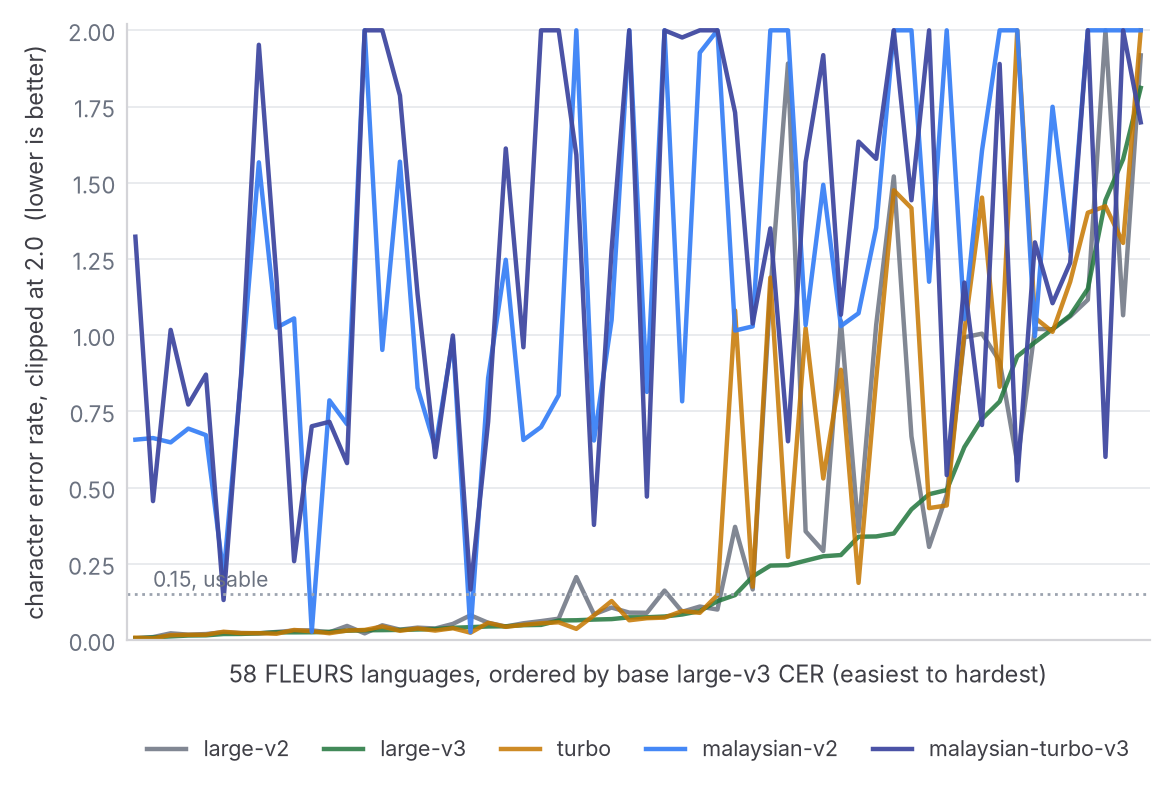}
  \caption{The same measurement language by language, ordered by base \texttt{large-v3}
  difficulty. Whisper genuinely cannot transcribe the right hand end: Amharic, Khmer and Somali
  sit above 1.4 for every checkpoint. The two Malay fine-tunes sit above the usable line almost
  everywhere, which is the damage LibriSpeech cannot report.}
  \label{fig:fleursperlang}
\end{figure}

Figure~\ref{fig:fleursperlang} plots it language by language, ordered by base difficulty, so
the languages Whisper cannot do at all are separated from the ones a fine-tune broke.

\texttt{malaysian-turbo-v3} looks broken in Table~\ref{tab:wer} and is not. Drop the roughly 1\% of
clips carrying a token run of six or more and \texttt{genuine} falls from 60.7\% to 33.9\%, and
\texttt{speech\_in\_noise} from 58.7\% to 25.6\%, the best in the table. One percent of clips
carries 27 points of word error rate. Looping is also what separates \texttt{turbo} from
\texttt{large-v3} on FLEURS, 4.7\% of clips against 0.9\%.

\subsection{The positive arm}

\begin{figure}[H]
  \centering
  \includegraphics[width=0.78\linewidth]{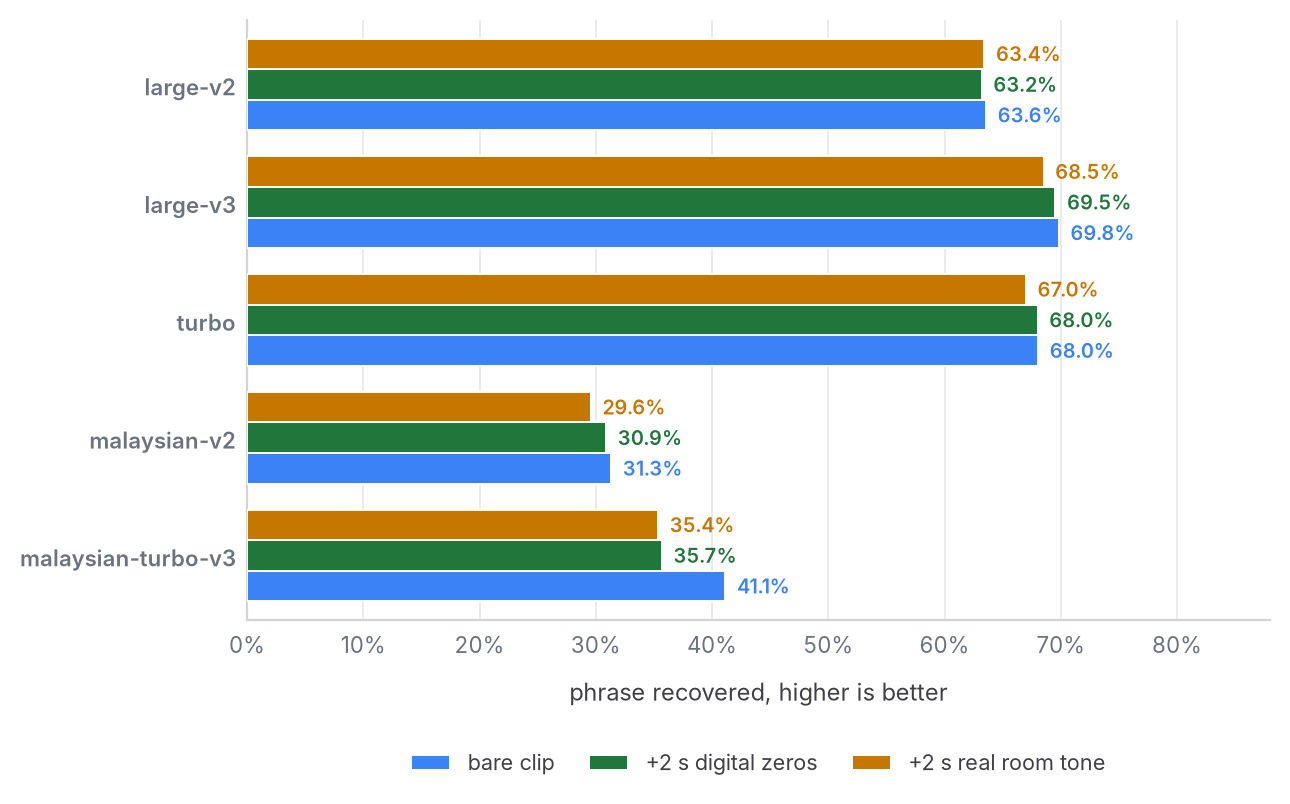}
  \caption{The ranking inverts when the phrase is really spoken. 6,267 clips in 83 languages, each run three ways. Padding is applied to the same clips, so each triple is a within-clip comparison.}
  \label{fig:lexsynth}
\end{figure}

Figure~\ref{fig:lexsynth} shows all three framings of the same clips, so the effect of
padding can be read within a checkpoint rather than across them.

Table~\ref{tab:lexsynth} gives the numbers behind that figure, including the mean, which is
worth reporting only next to the median.

\begin{table}[!htbp]
\centering
\caption{\texttt{lexicon\_synth} test split. Emitting the phrase is correct here.}
\label{tab:lexsynth}
\footnotesize
\begin{tabular}{lllrl}
\toprule
\schead \scth{model} & \scth{recovered, bare $\to$ tone} & \scth{median CER} & \scth{mean CER} & \scth{$>2\times$} \\
\midrule
\texttt{whisper-large-v2} & 63.6\% to 63.4\% & 0.125 & 0.378 to 0.384 & 0.6\% to 0.7\% \\
\textbf{\texttt{whisper-large-v3}} & \textbf{69.8\% to 68.5\%} & \textbf{0.089} & 0.288 to 0.302 & 0.2\% to 0.4\% \\
\texttt{whisper-large-v3-turbo} & 68.0\% to 67.0\% & 0.087 & 0.348 to 0.382 & 0.7\% to 0.9\% \\
\texttt{malaysian-whisper-v2} & 31.3\% to 29.6\% & 0.800 & 0.770 to \textbf{1.151} & 2.6\% to 3.7\% \\
\texttt{Malaysian-turbo-v3} & 41.1\% to 35.4\% & 0.385 & 1.508 to \textbf{3.749} & 3.0\% to 5.7\% \\
\bottomrule
\end{tabular}
\end{table}

\texttt{Malaysian-turbo-v3} wins every non-speech arm in Table~\ref{tab:nonspeech} and recovers
41.1\% of genuinely spoken phrases, against 69.8\% for \texttt{large-v3}
(Figure~\ref{fig:lexsynth}). Its silence is a prior
against emitting text, not a detector. Report one side only and you will ship it.

\begin{figure}[H]
  \centering
  \includegraphics[width=0.78\linewidth]{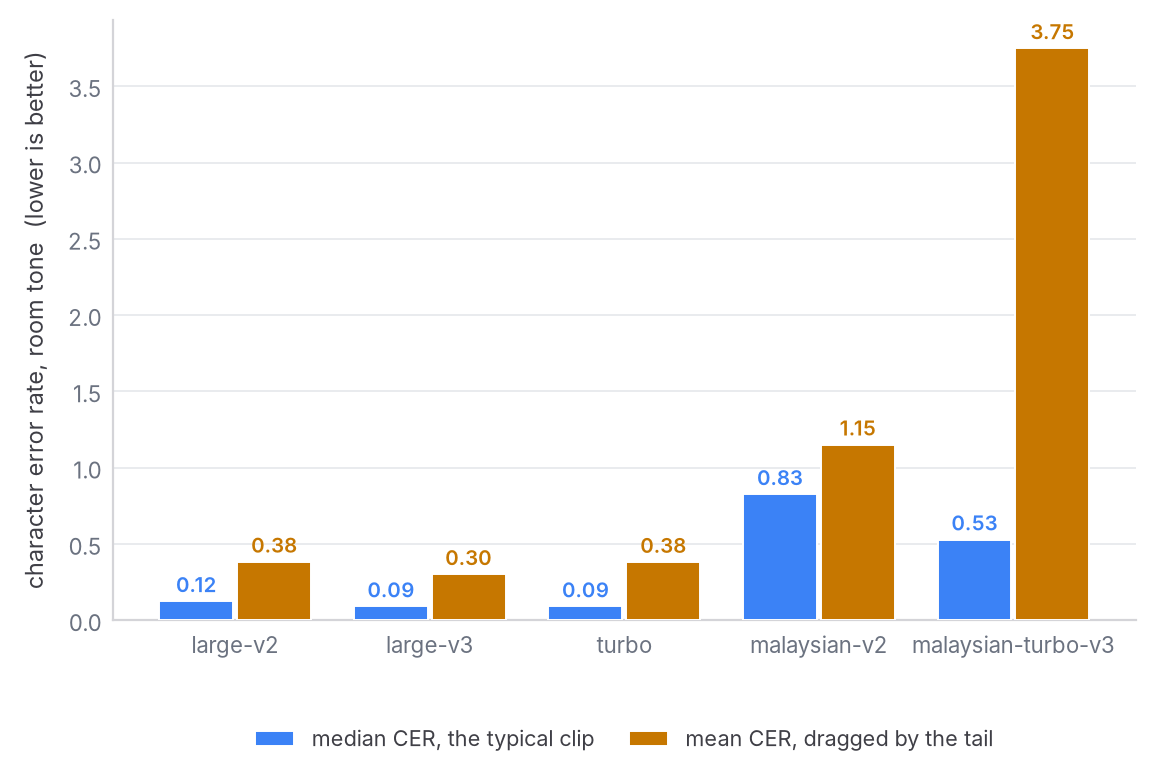}
  \caption{Mean character error rate on the positive arm is a tail statistic. The median barely
  moves between checkpoints while the mean triples, because a few percent of clips run past
  twice the phrase length. Report both.}
  \label{fig:lexsynthtail}
\end{figure}

Room tone costs the fine-tunes and not the base models. Same speech, two seconds of noise floor
on each side: \texttt{malaysian-v2} gains 0.38 mean character error rate,
\texttt{malaysian-turbo-v3} gains 2.24, \texttt{large-v3} gains 0.014. Mean character error rate here is
a tail statistic. The median moves by hundredths while the mean triples, because 3\% to 6\% of clips run past
twice the phrase length (Figure~\ref{fig:lexsynthtail}).

\subsection{Results on wild audio}

\begin{figure}[H]
  \centering
  \includegraphics[width=0.78\linewidth]{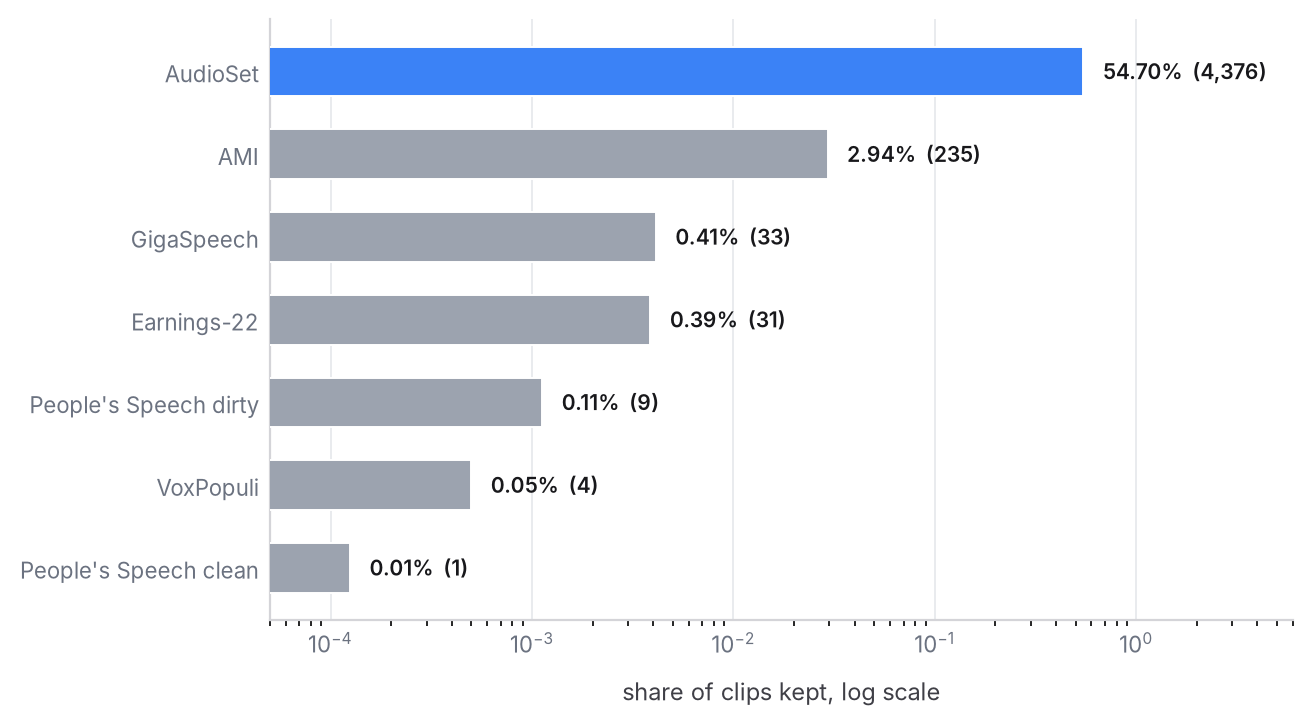}
  \caption{Hallucination yield by corpus, per 8,000 clips heard and mined with \texttt{large-v3}, on a log scale. Curated corpora barely fail. GigaSpeech is the right material and still yields 0.41\%, because its segments are cut to speech.}
  \label{fig:wildyield}
\end{figure}

Figure~\ref{fig:wildyield} plots the yield by corpus on a log scale, which is the only way
to show a range this wide on one axis.

\begin{figure}[H]
  \centering
  \includegraphics[width=0.78\linewidth]{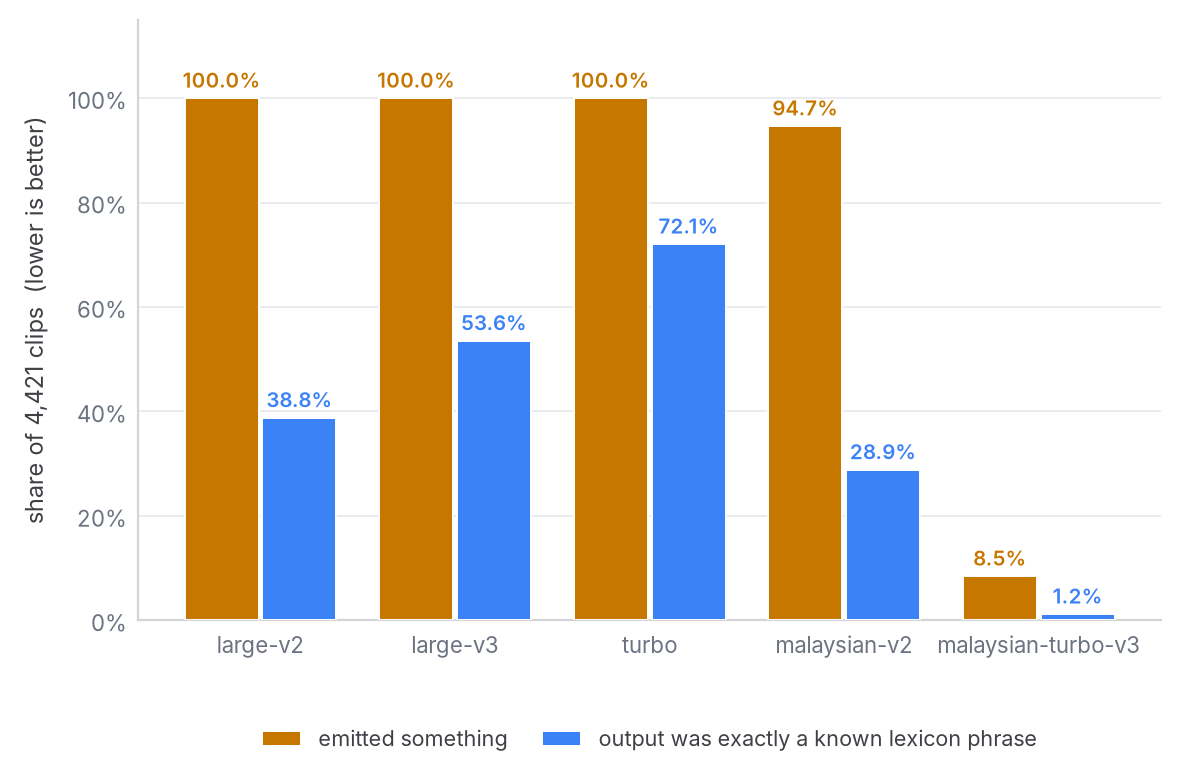}
  \caption{Real audio that a voice activity detector confirms has no speech in it. Every OpenAI
  checkpoint writes something on all 4,421 clips. Most of what they write is a phrase from the
  lexicon, which is what makes a blocklist work, but only once the clip is already known to be
  blank.}
  \label{fig:wildblank}
\end{figure}

On 4,421 clips that a voice activity detector confirms contain no speech, every OpenAI
checkpoint emits words on 100\% of them (Figure~\ref{fig:wildblank}). \texttt{Malaysian-turbo-v3} emits words on 8.5\%. Of
those outputs, 53.6\% for \texttt{large-v3} and 72.1\% for \texttt{turbo} are exactly a phrase
from the lexicon, against roughly 18\% to 22\% on speech clips from the same corpus. The
blocklist works, but only once the audio is already known to be blank.

Most wild loops are model-specific. On clips mined because \texttt{large-v3} looped,
\texttt{turbo} loops on 17.2\%. A loop corpus mined with one model is not a fair loop benchmark
for another.

\section{Lexicon}
\label{sec:lexicon}

Hallucinations repeat themselves. \texttt{thank you for watching}, \texttt{so}, the Russian
\texttt{Prodolzhenie sleduet...} (``to be continued''), and the Malay
\texttt{terima kasih kerana menonton} appear over and over on audio containing none of them. We merged four public sources into 40,891 phrases across 100 languages
(Table~\ref{tab:lexsources}).

\begin{table}[!htbp]
\centering
\caption{Where the lexicon comes from. All four are public and permissively licensed, and all
four are vendored in the release so the merge is reproducible.}
\label{tab:lexsources}
\footnotesize
\begin{tabular}{L{0.34\linewidth}L{0.26\linewidth}rl}
\toprule
\schead \scth{source} & \scth{what it is} & \scth{phrases} & \scth{licence} \\
\midrule
AGH DSP full tally \citep{baranski2025nonspeech} & Whisper output over non-speech audio & 30{,}407 & MIT \\
AGH DSP ``Bag of Hallucinations'' \citep{baranski2025nonspeech} & the curated subset of the above & 294 & MIT \\
\texttt{whisper-hallucinations} \citep{arbonel2024hfhalluc} & community-collected, multilingual & 7{,}889 & MIT \\
NVIDIA NeMo SDP, Granary \citep{koluguri2025granary} & the pipeline's own filter lists, 23 languages & 3{,}028 & Apache-2.0 \\
\bottomrule
\end{tabular}
\end{table}

The fourth row is worth dwelling on. Those files are the lists NVIDIA's pipeline uses to
\emph{delete} hallucinated pseudo-labels from Granary, shipped alongside the
\texttt{DetectWhisper\allowbreak Hallucination\allowbreak Features} processor that applies them. We use the same lists
to \emph{generate} training data. The phrases a corpus builder throws away are exactly the
phrases a model needs to be taught to handle.

The lexicon is the bridge between the two halves of this work. Read one way it is a blocklist.
Read the other way it is a list of sentences the model must still be able to transcribe when
somebody genuinely says them, which is exactly the training signal that is missing.

It needs filtering before use. It contains Whisper's own loop artefacts captured as phrases.
One entry is a single Gujarati character repeated seven times. These wreck character error rate
comparisons, one of them produced a CER of 22.9. We drop any entry that is a single repeated
token or has two or fewer distinct characters.

It also needs a frequency cut. Of 39,476 queued entries, 35,594 were observed exactly once, and
27,974 of those run five words or longer. They are raw recognition fragments, not phrases:
\texttt{0073a this vehicle s got 72 00 miles on it 3 5l v engine}. Synthesising those is
pointless, because the round trip fails by construction on digits and truncation. Two
independent text-to-speech systems score character error rate near 1.0 on them, which is the
signal that the input is at fault rather than the generator. We synthesise entries observed at
least twice: 3,882 entries across 85 languages, median two words.

\section{Synthesis}

To turn the lexicon into training data we need each phrase spoken aloud, in its own language, in
more than one voice. That is a text-to-speech problem, and the choice of system matters more
than it looks.

\subsection{Choosing a generator}

\begin{figure}[H]
  \centering
  \includegraphics[width=0.78\linewidth]{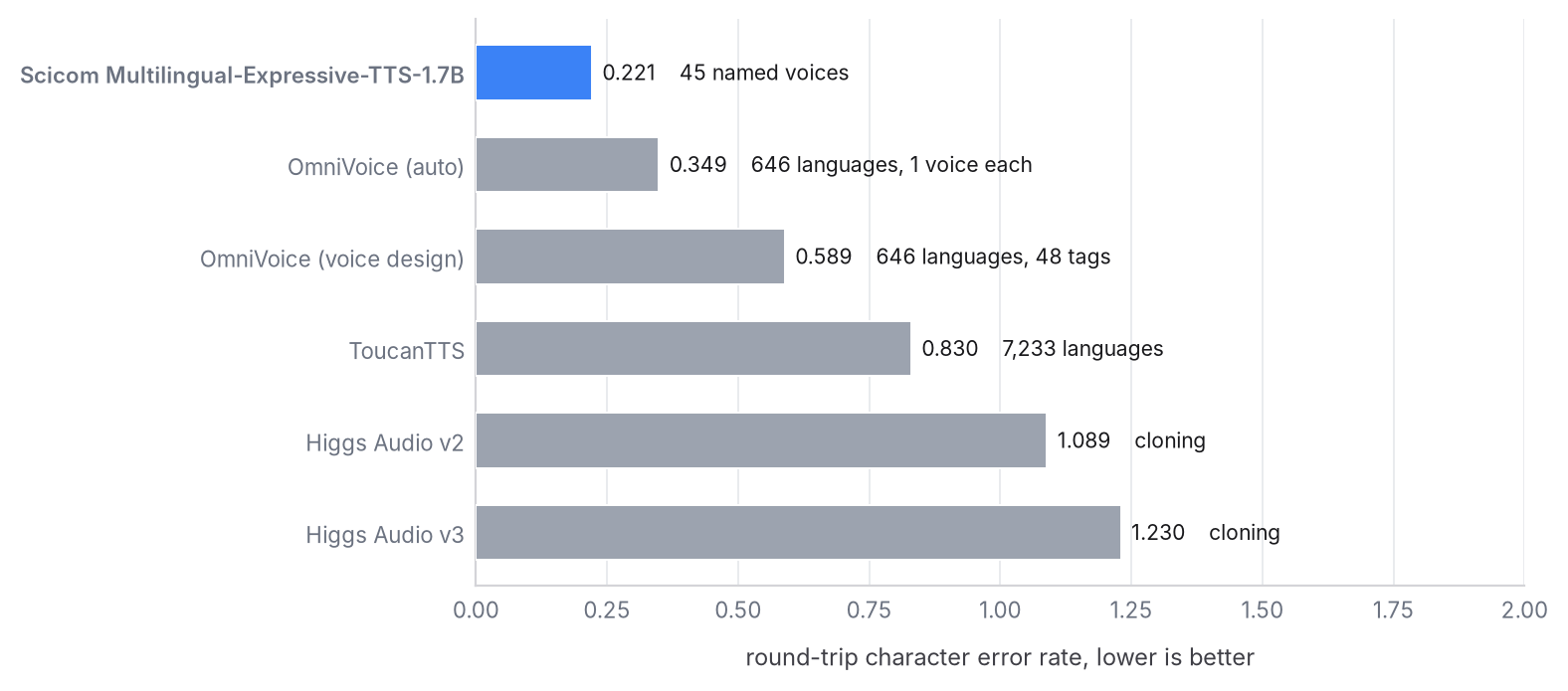}
  \caption{Text-to-speech candidates on 48 lexicon phrases in 22 languages, identical text, scored by recognition round trip. Coverage and quality are not the same axis. Scicom's own model is the blue bar.}
  \label{fig:tts}
\end{figure}

Figure~\ref{fig:tts} ranks the six candidates on identical text, with the number of
languages each covers beside its bar, because the two do not move together.

Table~\ref{tab:tts} gives the same ranking with the medians and the per-language win counts,
which matter because one bad language can carry a mean.

\begin{table}[!htbp]
\centering
\caption{Text-to-speech candidates. Round-trip character error rate, lower is better.}
\label{tab:tts}
\footnotesize
\begin{tabular}{L{0.32\linewidth}rrrll}
\toprule
\schead & \scth{mean CER} & \scth{median} & \scth{wins} & \scth{languages} & \scth{voices} \\
\midrule
\textbf{Scicom} Multilingual-Expressive-TTS-1.7B & \textbf{0.221} & \textbf{0.000} & \textbf{16} & untagged & 45 named \\
OmniVoice (auto) & 0.349 & 0.018 & 2 & \textbf{646} & \textbf{1 per language} \\
OmniVoice (voice design) & 0.589 & 0.062 & 3 & \textbf{646} & 48 tags \\
ToucanTTS & 0.830 & 0.586 & 0 & \textbf{7{,}233} & sampled from a GAN \\
Higgs Audio v2 & 1.089 & 0.713 & 0 & untagged & cloning \\
Higgs Audio v3 & 1.230 & 0.483 & 1 & 82/100 & cloning \\
\bottomrule
\end{tabular}
\end{table}

Coverage and quality are not the same axis. ToucanTTS covers 7,233 languages and wins nothing.
OmniVoice covers 646 and is the only system whose coverage is enumerable, so it drives the
hundred-language sweep, with Scicom's own Multilingual-Expressive-TTS-1.7B handling Malay,
English, Chinese and Tamil. It is the only model in the comparison that we built, and it is
marked as such in Table~\ref{tab:tts} and Figure~\ref{fig:tts} so the reader can discount it
if they wish. Its 0.221 is measured in speaker-name mode, which is the only mode it is good in;
its zero-shot cloning path is the worst system in this paper and Section~\ref{sec:vc} shows
why.

\subsection{Voice diversity}

\begin{figure}[H]
  \centering
  \includegraphics[width=0.78\linewidth]{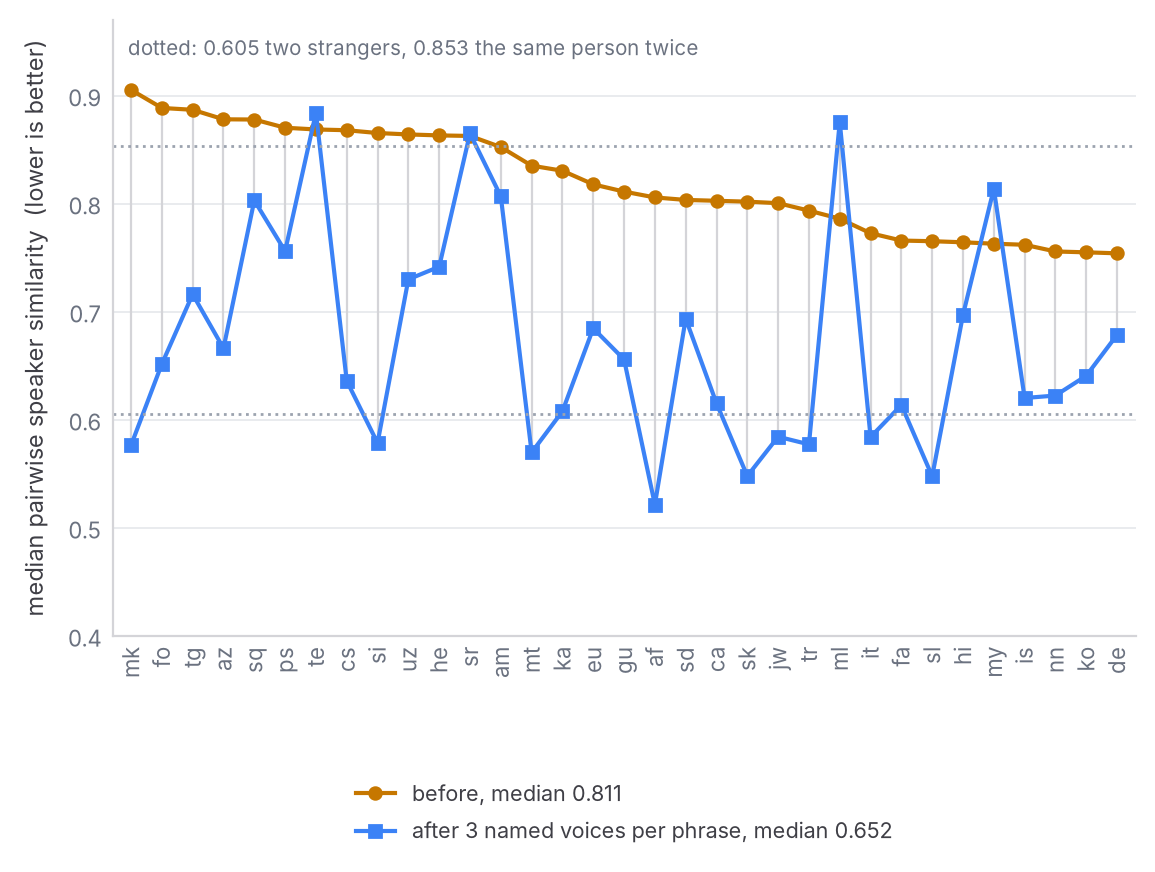}
  \caption{One voice per language, and what fixed it, on the 33 OmniVoice languages that sat at 0.75 or worse. Pairwise speaker similarity on a calibrated scale where 0.605 is two strangers and 0.853 is two clips of the same person.}
  \label{fig:diversity}
\end{figure}

Figure~\ref{fig:diversity} plots the 33 collapsed languages before and after the top-up, on
the calibrated scale where 0.605 is two strangers and 0.853 is one person twice.

OmniVoice takes no speaker argument, so every clip it produces carries an empty voice field. We
measured what that means with WavLM speaker embeddings \citep{chen2021wavlm} on a calibrated
scale: 33 of its 74 languages sat at a median pairwise cosine of 0.75 or worse
(Figure~\ref{fig:diversity}). One voice per language. A corpus built that way teaches the model one speaker's phrase, not the phrase.

The obvious fix, reference cloning, is the wrong one (Table~\ref{tab:diversity}).

\begin{table}[!htbp]
\centering
\caption{Two routes to distinct voices, probed on the same 4 languages, 24 phrases and 4 voices.
Lower cosine means better separated voices.}
\label{tab:diversity}
\footnotesize
\begin{tabular}{L{0.30\linewidth}lll}
\toprule
\schead \scth{route} & \scth{yield mk / gu / it} & \scth{median CER} & \scth{different-voice cosine} \\
\midrule
\textbf{Scicom Multilingual-Expressive, speaker name} & \textbf{86\% / 80\% / 79\%} & 0.07 / 0.17 / 0.00 & \textbf{0.65 / 0.64 / 0.58} \\
OmniVoice, reference cloning & 46\% / 68\% / 53\% & 0.32 / 0.25 / 0.26 & 0.68 / 0.72 / 0.69 \\
auto mode, for reference & 89\% / 73\% / 83\% & 0.02 / 0.11 / 0.00 & 0.91 / 0.81 / 0.77 \\
\bottomrule
\end{tabular}
\end{table}

Cloning costs about half the yield and separates the voices less well than simply naming a
speaker. So we topped up 26 languages rather than replacing them: three named voices per phrase,
auto-mode clips kept. 12,190 of 15,675 new clips passed the quality gate, 77.8\%, and the median
collapsed language moved from 0.81 to 0.63.

\subsection{Voice conversion}
\label{sec:vc}

\begin{figure}[H]
  \centering
  \includegraphics[width=0.78\linewidth]{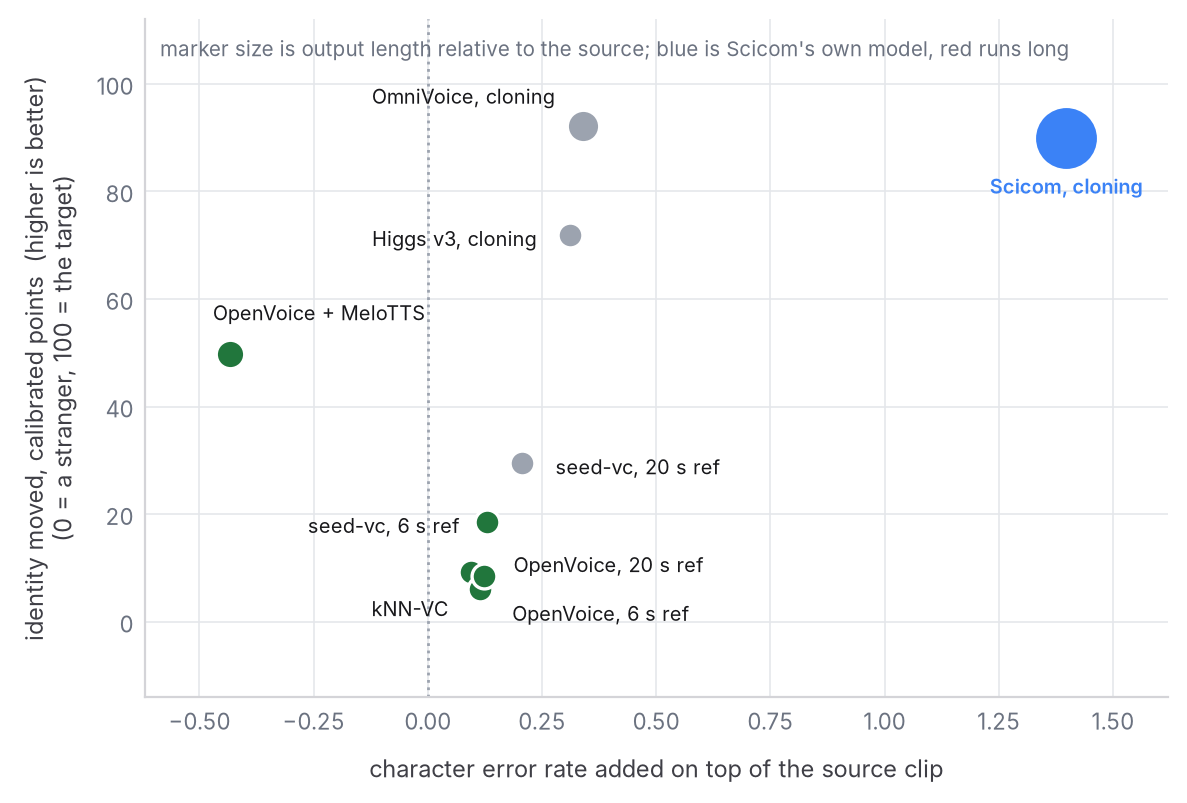}
  \caption{Voice conversion candidates. Character error rate added on top of the source clip against how far the voice actually moved. A converter that returns its input scores a perfect zero cost and is worthless, which is why the vertical axis is here at all.}
  \label{fig:vc}
\end{figure}

Figure~\ref{fig:vc} plots what each converter costs against how far it actually moved the
voice, with marker size for output length, so a system that runs long cannot hide behind a
good similarity score.

Character error rate alone picks the wrong winner here, because a converter that returns its
input unchanged scores a perfect zero degradation and is worthless. We report the cost added on
top of the source clip, similarity to the target and to the source on the same calibrated
scale, and the gap between them, which is the conversion that actually happened
(Figure~\ref{fig:vc}).

\textbf{Our own model is only usable in one of its two modes.}
Multilingual-Expressive can be conditioned either on a speaker name from its training
inventory or on a reference clip. The two land at opposite extremes on the same 184 clips in
the same 21 languages (Table~\ref{tab:twomodes}).

\begin{table}[!htbp]
\centering
\caption{The two ways to condition Scicom's Multilingual-Expressive, measured on the same
phrases. Naming a speaker is the best result in the whole comparison. Zero-shot cloning from a
reference clip is the worst, and the reason is duration, not timbre.}
\label{tab:twomodes}
\footnotesize
\begin{tabular}{L{0.26\linewidth}rrrrr}
\toprule
\schead \scth{conditioning} & \scth{CER} & \scth{$\Delta$CER} & \scth{$\to$ target} & \scth{duration $\times$} & \scth{$>2\times$} \\
\midrule
speaker name & \textbf{0.308} & \textbf{$-$0.035} & n/a & \textbf{1.03} & \textbf{2.7\%} \\
reference clip, zero shot & 1.740 & $+$1.397 & 0.861 & 3.27 & 67.9\% \\
\addlinespace[2pt]
\textit{Higgs v3, for comparison} & \textit{0.653} & \textit{$+$0.311} & \textit{0.813} & \textit{0.90} & \textit{2.7\%} \\
\bottomrule
\end{tabular}
\end{table}

In speaker-name mode it is the intelligibility winner outright, and the only system anywhere in
this comparison with a negative $\Delta$CER: the re-voiced clip is easier to recognise than the
OmniVoice clip it was asked to re-voice, across all 21 languages, at the right length.

Zero shot from a reference clip it is the worst system we measured. It matches the target
voice better than anything else, 0.861 raw cosine against Higgs v3's 0.813, so the timbre is
not the problem. It does not stop. The median output is \textbf{3.27 times} the length of the
source and \textbf{67.9\%} of clips run past twice, which is what drags character error rate to
1.740. On two-word phrases it keeps talking. Read the similarity with that in mind as well,
because a longer clip gives a better speaker-embedding estimate, so the voice match is
flattered by the same over-generation that ruins the transcript.

The practical consequence for this paper is that the positive pool is built with speaker names
only. Anyone reaching for this model to clone a specific voice from one reference clip should
use Higgs v3 or OmniVoice instead, or fix the duration first with a stop condition or by
trimming to the phrase.

\subsection{Quality gate}

Every synthesised clip passes a recognition round trip: the recogniser must recover the phrase,
and the clip must run about as long as the phrase should take. The gate is per language and
anchored to the judge's own floor, which is Whisper's character error rate on real FLEURS speech
in that language. A language where the judge itself scores 0.88, as it does on Burmese, cannot
be held to a 0.25 gate. Nine such languages are excluded rather than scored.

The final corpus is 29,112 clips in 83 languages, 22,845 train and 6,267 test, 15.8 hours, 92\%
non-English.

\section{Training sweep}
\label{sec:sweep}

\subsection{Setup}

We fine-tune \texttt{whisper-large-v3} with LoRA \citep{hu2021lora} and with full fine-tuning.
The target for a blank clip is the empty transcript under that clip's language token, which is
what teaches ``no words here'' rather than ``emit something short''. The four prompt tokens are
masked out of the loss, so loss falls only on the transcript and the final end-of-text
token. Table~\ref{tab:trainsets} lists what goes into the mix, how much of it there is, and
what each part is there to teach.

\begin{table}[!htbp]
\centering
\caption{Training sets. Everything is disjoint from the benchmark by construction and every
test split stays held out.}
\label{tab:trainsets}
\footnotesize
\begin{tabular}{L{0.26\linewidth}rrlL{0.28\linewidth}}
\toprule
\schead \scth{source} & \scth{clips} & \scth{hours} & \scth{target} & \scth{teaches} \\
\midrule
\texttt{corpus} nonspeech (FSD50K \texttt{dev}) & 16{,}983 & 33.3 & empty & noise means say nothing \\
\texttt{corpus} music (FMA shards 2-12) & 7{,}170 & 58.3 & empty & music means say nothing \\
\texttt{corpus} Malay speech, Emilia \citep{he2024emilia} & 5{,}801 & 14.8 & transcript & keep accuracy \\
\texttt{corpus} reduplication & 2{,}514 & 4.0 & exact repeats & do not run away \\
\texttt{corpus} silence & 59 & 0.7 & empty & room tone means say nothing \\
\texttt{lexicon\_synth} train & 22{,}845 & 12.5 & the phrase & the phrase is real, transcribe it \\
\texttt{wild} train & 3{,}656 & 9.6 & empty & real noisy audio means say nothing \\
\bottomrule
\end{tabular}
\end{table}

\subsection{Mixes}

The sweep answers a paired question, and both halves are run at every configuration.

\begin{enumerate}
\item Without the synthetic lexicon, does fine-tuning improve hallucination and repetition on
real audio?
\item With the synthetic lexicon, does it improve them further, and what does it cost?
\end{enumerate}

Two mixes, identical except for the synthetic positives (Table~\ref{tab:mixes}).

\begin{table}[!htbp]
\centering
\caption{The paired mixes. \texttt{no\_synth} is \texttt{all} with \texttt{lexicon\_synth}
removed and nothing else changed.}
\label{tab:mixes}
\footnotesize
\begin{tabular}{lrrl}
\toprule
\schead \scth{mix} & \scth{clips} & \scth{blank share} & \scth{contains \NoCaseChange{lexicon\_synth}} \\
\midrule
\texttt{no\_synth} & 36{,}183 & 77\% & no \\
\texttt{all} & 59{,}028 & 47\% & yes, 22{,}845 clips \\
\bottomrule
\end{tabular}
\end{table}

Removing the synthetic positives also moves the blank share from 47\% to 77\%. That is not a
confound to be stripped out, it is the mechanism. The synthetic positives are the counterweight,
and removing them necessarily makes the mix more blank-dominated.

\subsection{Grid}

LoRA alpha tracks twice the rank, so rank is the only thing that varies within a family. Full fine-tunes use batch 4 with 4 accumulation steps to hold the same 16 clips per optimiser
step as LoRA's 8 with 2, so effective batch size does not confound the comparison.
Table~\ref{tab:trainconfig} lists everything else, and none of it varies across the 66 runs.

\begin{table}[!htbp]
\centering
\caption{Training configuration, identical for every one of the 66 runs except the three
factors being swept: method, rank and learning rate. One GPU per run, so the per-device batch
is the batch.}
\label{tab:trainconfig}
\footnotesize
\begin{tabular}{L{0.26\linewidth}L{0.58\linewidth}}
\toprule
\schead \scth{setting} & \scth{value} \\
\midrule
optimiser & AdamW, fused \\
$\beta_1$, $\beta_2$, $\epsilon$ & 0.9, 0.999, $1\times10^{-8}$ \\
weight decay & 0.0 \\
schedule & linear decay, 50 warmup steps \\
gradient clipping & 1.0 \\
precision & bf16 \\
batch $\times$ accumulation, LoRA & $8 \times 2 = 16$ clips per step \\
batch $\times$ accumulation, full & $4 \times 4 = 16$ clips per step \\
steps & 1{,}000, so 16{,}000 clips seen \\
max label length & 200 tokens \\
loss & on the transcript and the final end-of-text token only; the four prompt tokens are masked to $-100$ \\
seed & 0 \\
\bottomrule
\end{tabular}
\end{table}

\textbf{Which modules the adapters touch is one of the swept factors.} A Whisper block has four
attention projections and two feed-forward matrices, and the feed-forward pair is the larger
half. \texttt{fc1} maps 1{,}280 to 5{,}120 and \texttt{fc2} maps back, so each of them takes
two and a half times the low-rank parameters of a $1{,}280 \times 1{,}280$ attention
projection at the same rank. We sweep both choices rather than assuming one
(Table~\ref{tab:targets}).

\begin{table}[!htbp]
\centering
\caption{The two adapter families. Trainable counts are measured from the runs, not estimated,
and are exactly linear in rank. At every rank, adding the MLP pair costs 1.83 times the
attention-only count.}
\label{tab:targets}
\footnotesize
\begin{tabular}{llrrrrr}
\toprule
\schead \scth{family} & \scth{modules} & \scth{$r = 8$} & \scth{$r = 16$} & \scth{$r = 32$} & \scth{$r = 64$} & \scth{$r = 128$} \\
\midrule
attention only & \texttt{q\_proj}, \texttt{k\_proj}, \texttt{v\_proj}, \texttt{out\_proj}
 & 7.9\,M & 15.7\,M & 31.5\,M & 62.9\,M & 125.8\,M \\
 & & 0.51\% & 1.01\% & 2.00\% & 3.92\% & 7.54\% \\
\addlinespace[2pt]
attention + MLP & the four above, plus \texttt{fc1}, \texttt{fc2}
 & 14.4\,M & 28.8\,M & 57.7\,M & 115.3\,M & 230.7\,M \\
 & & 0.93\% & 1.83\% & 3.60\% & 6.95\% & 13.00\% \\
\bottomrule
\end{tabular}
\end{table}

That is five ranks in each of the two families at three learning rates, thirty adapter
configurations, plus three full fine-tunes: 33 per mix and 66 runs in total. Both mixes
throughout, so every configuration is a matched pair (Table~\ref{tab:grid}).

\begin{table}[!htbp]
\centering
\caption{The grid. Every cell is run on both mixes. 1e-3 is absent on purpose, see below.}
\label{tab:grid}
\footnotesize
\begin{tabular}{lll}
\toprule
\schead \scth{method} & \scth{ranks ($\alpha = 2r$)} & \scth{learning rates} \\
\midrule
LoRA, attention only & 8, 16, 32, 64, 128 & 1e-4, 2e-4, 5e-4 \\
LoRA, attention + MLP & 8, 16, 32, 64, 128 & 1e-4, 2e-4, 5e-4 \\
full fine-tune & all 1{,}574.9 M parameters & 5e-6, 1e-5, 2e-5 \\
\bottomrule
\end{tabular}
\end{table}

Every run is trained for 1,000 steps with 50 warmup steps in bf16, then scored on the complete
benchmark: the eight published arms, the synthetic positives, the wild test split, and FLEURS.

\textbf{Control the optimiser before attributing anything to the data.} Our first sweep ran at
1e-3. The \texttt{corpus} mix finished at loss 9.76 while mixes containing the same clips
finished at 1.1 to 1.2, which reads as evidence that a blank-heavy mix is bad. It is not. The
same mix at 2e-4 converges cleanly to 0.64. The checkpoint from the diverged run hallucinates on
100\% of silence and 100\% of music. That says the optimiser broke, and says nothing about
the data.

\subsection{Results}

Table~\ref{tab:paired} is the answer to both questions. Every row is one fine-tune of
\texttt{whisper-large-v3}, and the two rows in each pair differ only by the 22,845 synthetic
positives.

{\footnotesize\begin{longtable}{llrrrrr}
\caption{The paired sweep, all 33 configurations. \texttt{wild words} is the share of 4,421 real voice-free clips where the model emitted words, \texttt{wild loop} the share carrying a token run of six or more, \texttt{recovered} the share of genuinely spoken phrases it got back. Lower is better everywhere except \texttt{recovered}.}\label{tab:paired}\\
\toprule
\schead \scth{config} & \scth{mix} & \scth{wild words} & \scth{wild loop} & \scth{recovered} & \scth{ls WER} & \scth{FLEURS CER} \\
\midrule
\endfirsthead
\multicolumn{7}{l}{\footnotesize\itshape Table \thetable\ continued from the previous page}\\
\toprule
\schead \scth{config} & \scth{mix} & \scth{wild words} & \scth{wild loop} & \scth{recovered} & \scth{ls WER} & \scth{FLEURS CER} \\
\midrule
\endhead
\midrule \multicolumn{7}{r}{\footnotesize\itshape continued on the next page}\\
\endfoot
\bottomrule
\endlastfoot
\textit{base large-v3} & \textit{none} & 0.999 & 0.008 & 0.698 & 0.035 & 0.305 \\
\midrule
LoRA r8 attn+mlp, 1e-4 & \texttt{no\_synth} & 0.961 & 0.012 & 0.594 & 0.035 & 0.633 \\
 & \texttt{all} & 0.745 & 0.013 & 0.716 & 0.036 & 0.541 \\
\addlinespace[2pt]
LoRA r8 attn+mlp, 2e-4 & \texttt{no\_synth} & 0.969 & 0.011 & 0.587 & 0.034 & 0.603 \\
 & \texttt{all} & 0.758 & 0.013 & 0.762 & 0.035 & 0.507 \\
\addlinespace[2pt]
LoRA r8 attn+mlp, 5e-4 & \texttt{no\_synth} & 0.968 & 0.008 & 0.595 & 0.035 & 0.530 \\
 & \texttt{all} & 0.695 & 0.009 & 0.790 & 0.034 & 0.449 \\
\addlinespace[2pt]
LoRA r16 attn+mlp, 1e-4 & \texttt{no\_synth} & 0.961 & 0.009 & 0.577 & 0.035 & 0.657 \\
 & \texttt{all} & 0.742 & 0.012 & 0.732 & 0.036 & 0.547 \\
\addlinespace[2pt]
LoRA r16 attn+mlp, 2e-4 & \texttt{no\_synth} & 0.964 & 0.011 & 0.589 & 0.034 & 0.616 \\
 & \texttt{all} & 0.710 & 0.010 & 0.778 & \textbf{0.033} & 0.449 \\
\addlinespace[2pt]
LoRA r16 attn+mlp, 5e-4 & \texttt{no\_synth} & 0.979 & 0.005 & 0.585 & 0.036 & 0.537 \\
 & \texttt{all} & \textbf{0.611} & 0.004 & 0.810 & 0.035 & 0.385 \\
\addlinespace[2pt]
LoRA r32 attn+mlp, 1e-4 & \texttt{no\_synth} & 0.958 & 0.008 & 0.571 & 0.034 & 0.668 \\
 & \texttt{all} & 0.761 & 0.011 & 0.757 & 0.035 & 0.527 \\
\addlinespace[2pt]
LoRA r32 attn+mlp, 2e-4 & \texttt{no\_synth} & 0.948 & 0.008 & 0.602 & 0.034 & 0.549 \\
 & \texttt{all} & 0.619 & 0.008 & 0.799 & 0.035 & 0.419 \\
\addlinespace[2pt]
LoRA r32 attn+mlp, 5e-4 & \texttt{no\_synth} & 0.799 & 0.005 & 0.559 & 0.036 & 0.610 \\
 & \texttt{all} & 0.780 & 0.003 & 0.829 & 0.036 & 0.415 \\
\addlinespace[2pt]
LoRA r64 attn+mlp, 1e-4 & \texttt{no\_synth} & 0.956 & 0.007 & 0.597 & 0.035 & 0.604 \\
 & \texttt{all} & 0.754 & 0.009 & 0.783 & 0.034 & 0.460 \\
\addlinespace[2pt]
LoRA r64 attn+mlp, 2e-4 & \texttt{no\_synth} & 0.968 & 0.005 & 0.599 & 0.035 & 0.533 \\
 & \texttt{all} & 0.793 & 0.016 & 0.818 & 0.035 & 0.426 \\
\addlinespace[2pt]
LoRA r64 attn+mlp, 5e-4 & \texttt{no\_synth} & 0.150 & 0.000 & 0.511 & 0.041 & 0.680 \\
 & \texttt{all} & 0.478 & 0.005 & 0.827 & 0.041 & 0.398 \\
\addlinespace[2pt]
LoRA r128 attn+mlp, 1e-4 & \texttt{no\_synth} & 0.925 & 0.005 & 0.591 & 0.035 & 0.530 \\
 & \texttt{all} & 0.807 & 0.011 & 0.798 & 0.035 & 0.461 \\
\addlinespace[2pt]
LoRA r128 attn+mlp, 2e-4 & \texttt{no\_synth} & 0.823 & 0.002 & 0.574 & 0.036 & 0.522 \\
 & \texttt{all} & 0.721 & 0.000 & 0.840 & 0.037 & \textbf{0.381} \\
\addlinespace[2pt]
LoRA r128 attn+mlp, 5e-4 & \texttt{no\_synth} & 0.071 & 0.002 & 0.292 & 0.113 & 0.839 \\
 & \texttt{all} & 0.241 & 0.005 & 0.816 & 0.047 & 0.464 \\
\addlinespace[2pt]
LoRA r8 attn only, 1e-4 & \texttt{no\_synth} & 0.943 & 0.018 & 0.565 & 0.035 & 0.696 \\
 & \texttt{all} & 0.911 & 0.017 & 0.662 & 0.035 & 0.613 \\
\addlinespace[2pt]
LoRA r8 attn only, 2e-4 & \texttt{no\_synth} & 0.949 & 0.012 & 0.577 & 0.034 & 0.641 \\
 & \texttt{all} & 0.900 & 0.023 & 0.699 & 0.034 & 0.583 \\
\addlinespace[2pt]
LoRA r8 attn only, 5e-4 & \texttt{no\_synth} & 0.937 & 0.008 & 0.588 & 0.034 & 0.624 \\
 & \texttt{all} & 0.801 & 0.012 & 0.767 & 0.034 & 0.503 \\
\addlinespace[2pt]
LoRA r16 attn only, 1e-4 & \texttt{no\_synth} & 0.942 & 0.014 & 0.566 & 0.034 & 0.702 \\
 & \texttt{all} & 0.900 & 0.023 & 0.680 & 0.035 & 0.652 \\
\addlinespace[2pt]
LoRA r16 attn only, 2e-4 & \texttt{no\_synth} & 0.944 & 0.012 & 0.584 & 0.034 & 0.642 \\
 & \texttt{all} & 0.842 & 0.016 & 0.733 & 0.034 & 0.540 \\
\addlinespace[2pt]
LoRA r16 attn only, 5e-4 & \texttt{no\_synth} & 0.942 & 0.008 & 0.588 & 0.034 & 0.593 \\
 & \texttt{all} & 0.798 & 0.013 & 0.804 & 0.034 & 0.451 \\
\addlinespace[2pt]
LoRA r32 attn only, 1e-4 & \texttt{no\_synth} & 0.929 & 0.014 & 0.575 & 0.034 & 0.657 \\
 & \texttt{all} & 0.885 & 0.023 & 0.720 & 0.034 & 0.570 \\
\addlinespace[2pt]
LoRA r32 attn only, 2e-4 & \texttt{no\_synth} & 0.943 & 0.009 & 0.597 & 0.033 & 0.594 \\
 & \texttt{all} & 0.883 & 0.014 & 0.759 & 0.034 & 0.506 \\
\addlinespace[2pt]
LoRA r32 attn only, 5e-4 & \texttt{no\_synth} & 0.895 & 0.005 & 0.584 & 0.035 & 0.574 \\
 & \texttt{all} & 0.885 & 0.008 & 0.818 & 0.035 & 0.473 \\
\addlinespace[2pt]
LoRA r64 attn only, 1e-4 & \texttt{no\_synth} & 0.970 & 0.013 & 0.583 & 0.034 & 0.674 \\
 & \texttt{all} & 0.843 & 0.017 & 0.749 & 0.034 & 0.475 \\
\addlinespace[2pt]
LoRA r64 attn only, 2e-4 & \texttt{no\_synth} & 0.979 & 0.005 & 0.599 & 0.034 & 0.550 \\
 & \texttt{all} & 0.894 & 0.011 & 0.786 & 0.034 & 0.521 \\
\addlinespace[2pt]
LoRA r64 attn only, 5e-4 & \texttt{no\_synth} & 0.948 & 0.004 & 0.581 & 0.036 & 0.559 \\
 & \texttt{all} & 0.896 & 0.009 & \textbf{0.846} & 0.036 & 0.384 \\
\addlinespace[2pt]
LoRA r128 attn only, 1e-4 & \texttt{no\_synth} & 0.971 & 0.013 & 0.596 & 0.034 & 0.621 \\
 & \texttt{all} & 0.845 & 0.012 & 0.767 & 0.034 & 0.522 \\
\addlinespace[2pt]
LoRA r128 attn only, 2e-4 & \texttt{no\_synth} & 0.958 & 0.002 & 0.600 & 0.035 & 0.552 \\
 & \texttt{all} & 0.854 & 0.016 & 0.818 & 0.036 & 0.420 \\
\addlinespace[2pt]
LoRA r128 attn only, 5e-4 & \texttt{no\_synth} & 0.999 & 0.000 & 0.000 & 0.985 & 0.995 \\
 & \texttt{all} & 0.762 & 0.002 & 0.825 & 0.040 & 0.402 \\
\addlinespace[2pt]
full, 5e-6 & \texttt{no\_synth} & 0.931 & 0.013 & 0.535 & 0.035 & 0.797 \\
 & \texttt{all} & 0.918 & 0.017 & 0.504 & 0.035 & 0.886 \\
\addlinespace[2pt]
full, 1e-5 & \texttt{no\_synth} & 0.937 & 0.009 & 0.548 & 0.035 & 0.809 \\
 & \texttt{all} & 0.866 & 0.013 & 0.615 & 0.035 & 0.761 \\
\addlinespace[2pt]
full, 2e-5 & \texttt{no\_synth} & 0.922 & 0.005 & 0.598 & 0.034 & 0.539 \\
 & \texttt{all} & 0.728 & 0.007 & 0.724 & 0.036 & 0.545 \\
\addlinespace[2pt]
\end{longtable}
}

\begin{figure}[H]
  \centering
  \includegraphics[width=\linewidth]{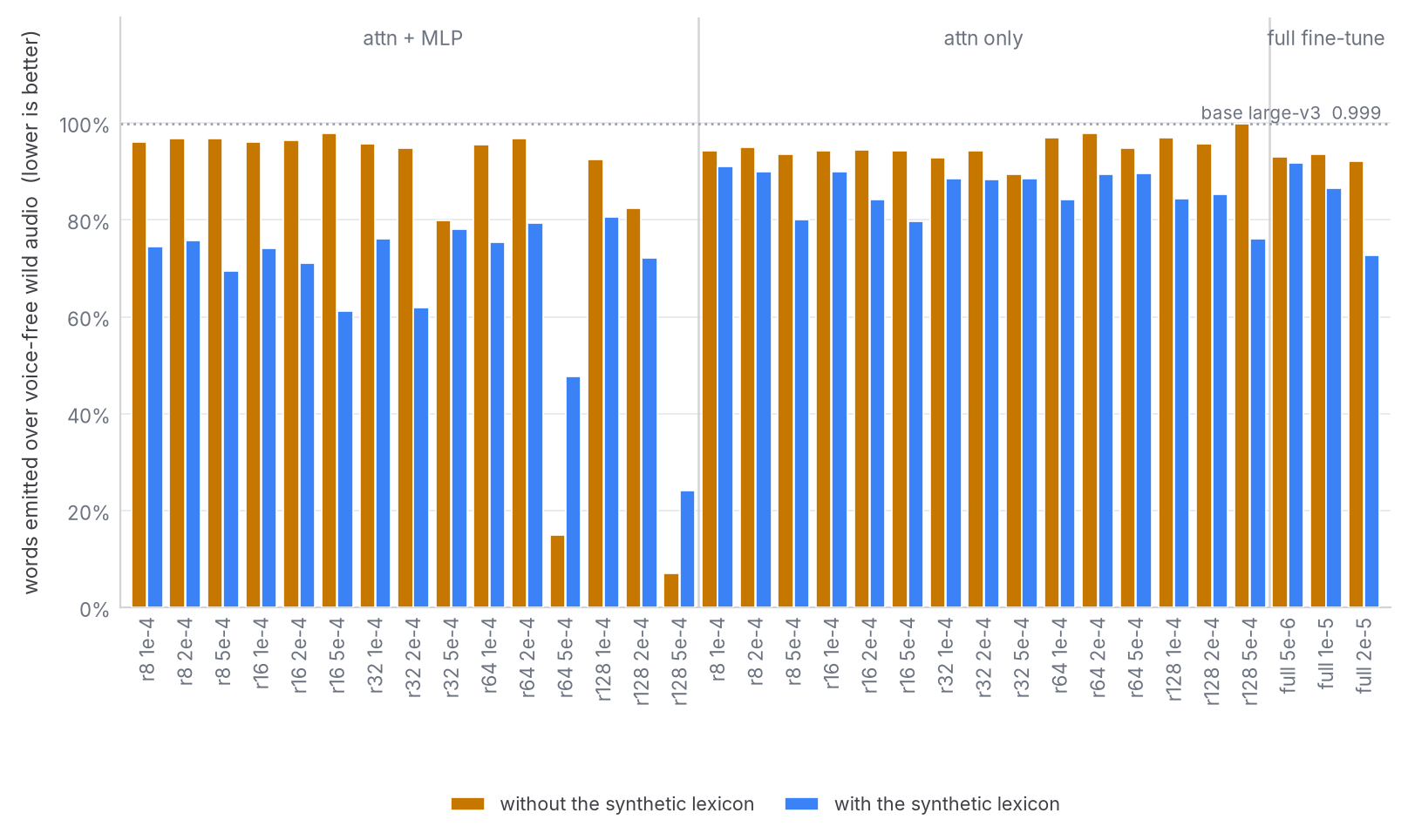}
  \caption{Hallucination on real voice-free audio, with and without the synthetic positives, at
  every configuration in the grid.}
  \label{fig:sweepwild}
\end{figure}

Figure~\ref{fig:sweepwild} puts all 33 configurations side by side, paired, so the effect of
the synthetic positives can be read off one bar against its neighbour.

\begin{figure}[H]
  \centering
  \includegraphics[width=0.78\linewidth]{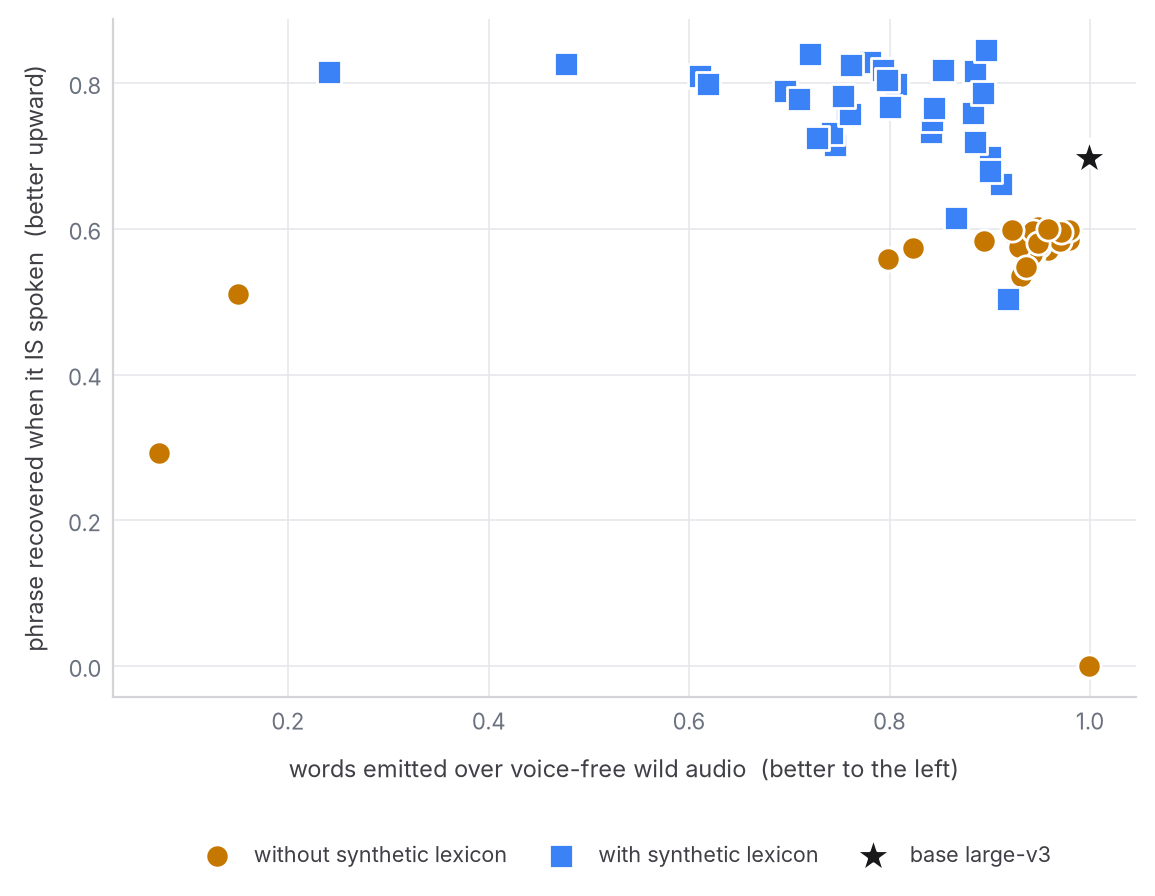}
  \caption{The trade from Figure~\ref{fig:tradeoff}, redrawn with the fine-tunes on it. The
  mix without the synthetic positives walks left and down. The mix with them walks left and
  up.}
  \label{fig:sweepcost}
\end{figure}

Figure~\ref{fig:sweepcost} redraws the trade-off of Figure~\ref{fig:tradeoff} with all 66
fine-tunes on it, which is the clearest way to see that the two mixes move in different
directions.

\textbf{Without the synthetic lexicon, hallucination falls a little and the model forgets a
lot.} Across the configurations that keep LibriSpeech word error rate at or under 0.040, word
emission on real voice-free audio moves from 0.999 to between 0.799 and 0.968. Phrase recovery
falls from 0.698 to between 0.511 and 0.602 in every one of them, and FLEURS character error
rate rises from 0.305 to between 0.522 and 0.809. That is the answer to the first question.
Fine-tuning on negatives alone does reduce hallucination, and it costs more accuracy than it
buys quiet.

Two configurations do silence the model. \texttt{no\_synth} at rank 64 and 5e-4 reaches 0.150,
and at rank 128 and 5e-4 it reaches 0.071, the quietest number anywhere in this paper. Look at
what came with it: recovery 0.511 and 0.292, FLEURS 0.680 and 0.839, and LibriSpeech word error
rate 0.113 at rank 128, three times the base model. That is \texttt{malaysian-turbo-v3}
reproduced from scratch in a thousand steps. A model can always be made quiet by making it
mute.

\textbf{With it, the model gets quieter and more accurate at the same time.} Across all 33
pairs, adding the synthetic positives lowers word emission on wild audio in 31, raises phrase
recovery in 32, and lowers FLEURS character error rate in 31. The median differences are
$-0.118$ on wild audio, $+0.186$ on recovery and $-0.107$ on FLEURS. LibriSpeech does not move:
the median difference is $0.000$, and neither mix wins it more often than the other.

Figure~\ref{fig:sweepwild} puts all 33 configurations side by side, paired, so the effect of
the synthetic positives can be read off one bar against its neighbour.
Restrict to the 30 pairs where both halves hold LibriSpeech word error rate at or under 0.040,
which is the only regime anybody would ship, and it is unanimous. \texttt{all} emits fewer
words on real voice-free audio in \textbf{30 of 30}, recovers more phrases in 29 of 30, and
holds multilingual accuracy better in 28 of 30. The configurations where \texttt{no\_synth}
looks quieter are the ones that failed the accuracy test.

The best single checkpoint in the grid is \texttt{all} at rank 64 and 5e-4: silence
hallucination 0.619 to \textbf{0.024}, words over wild voice-free audio 0.999 to
\textbf{0.478}, phrase recovery 0.698 to \textbf{0.827}, for LibriSpeech 0.035 to 0.041 and
FLEURS 0.305 to 0.398. It is quieter on noise and better on speech than the model it came from, which is the corner
Figure~\ref{fig:tradeoff} says is empty. Figure~\ref{fig:sweepcost} redraws that trade with
all 66 runs on it.

\begin{figure}[H]
  \centering
  \includegraphics[width=0.78\linewidth]{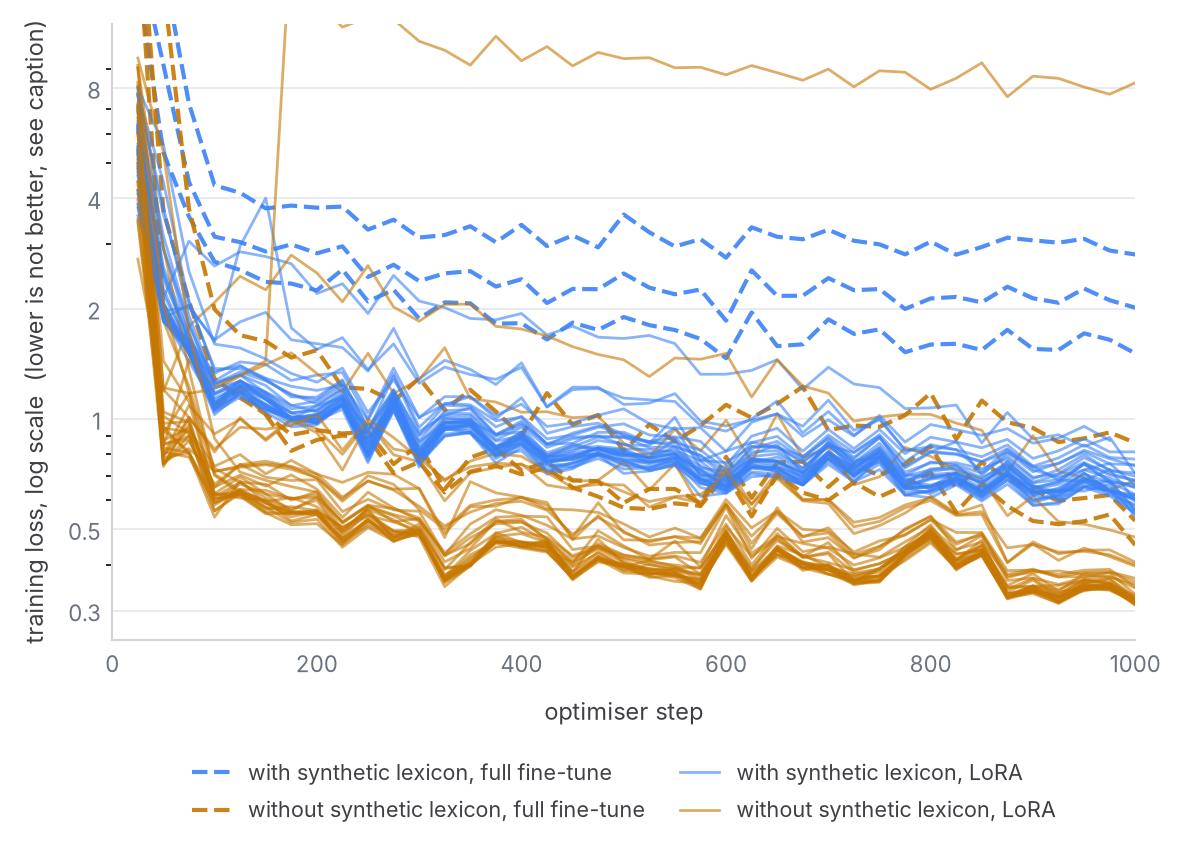}
  \caption{Training curves for all 66 runs. The mix without the synthetic positives sits below
  the mix with them for the whole of training, and it is the worse model on every axis that
  matters. An empty target is cheap to predict, and \texttt{no\_synth} is 77\% blank against
  \texttt{all} at 47\%, so the gap is arithmetic rather than evidence. Full fine-tunes (dashed)
  are still descending at step 1,000.}
  \label{fig:loss}
\end{figure}

Figure~\ref{fig:loss} shows all 66 training curves, coloured by mix, and the ordering in it
is the opposite of the ordering in every result that follows.

\begin{figure}[H]
  \centering
  \includegraphics[width=0.78\linewidth]{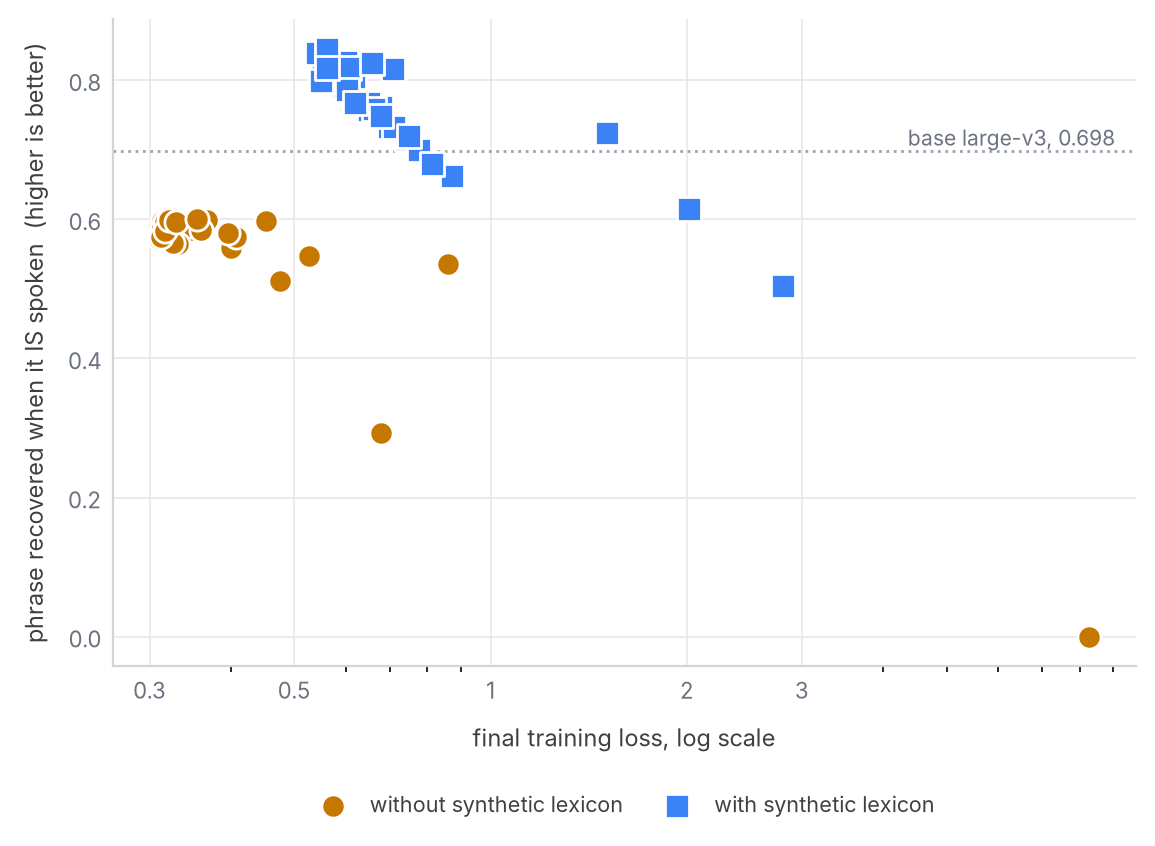}
  \caption{Final training loss against phrase recovery. The two mixes barely overlap on the
  loss axis and the ranking inverts across them, so loss cannot be used to choose between them.
  Within a mix it carries little signal either.}
  \label{fig:lossresult}
\end{figure}

Figure~\ref{fig:lossresult} plots final training loss against phrase recovery, which is the
cleanest way to show that the first predicts nothing useful about the second.

\textbf{Training loss is the wrong thing to watch here.} Twenty-eight of the thirty
\texttt{no\_synth} LoRA runs finish below \emph{every} \texttt{all} LoRA run. Median final
loss is 0.322 against 0.621. The mix that trains to a visibly better loss is the one that recovers less. Phrase recovery
separates the two completely: across the sixty LoRA runs, no checkpoint trained without the
positives exceeds 0.602 and none trained with them falls below 0.662 (Figure~\ref{fig:loss}, Figure~\ref{fig:lossresult}). The reason is arithmetic:
\texttt{no\_synth} is 77\% blank-labelled against \texttt{all} at 47\%, and predicting an empty
transcript is cheap. Comparing cross-entropy across mixes with different blank shares measures
the mixes, not the models. The full fine-tunes make the same point from the other side: they end at 1.51, 2.01 and 2.81
on \texttt{all} against 0.45, 0.53 and 0.86 on \texttt{no\_synth}, and all six are still
descending at step 1,000, which is what a thousand steps of 16 clips looks like on 1.5 B
parameters.

\begin{figure}[H]
  \centering
  \includegraphics[width=0.78\linewidth]{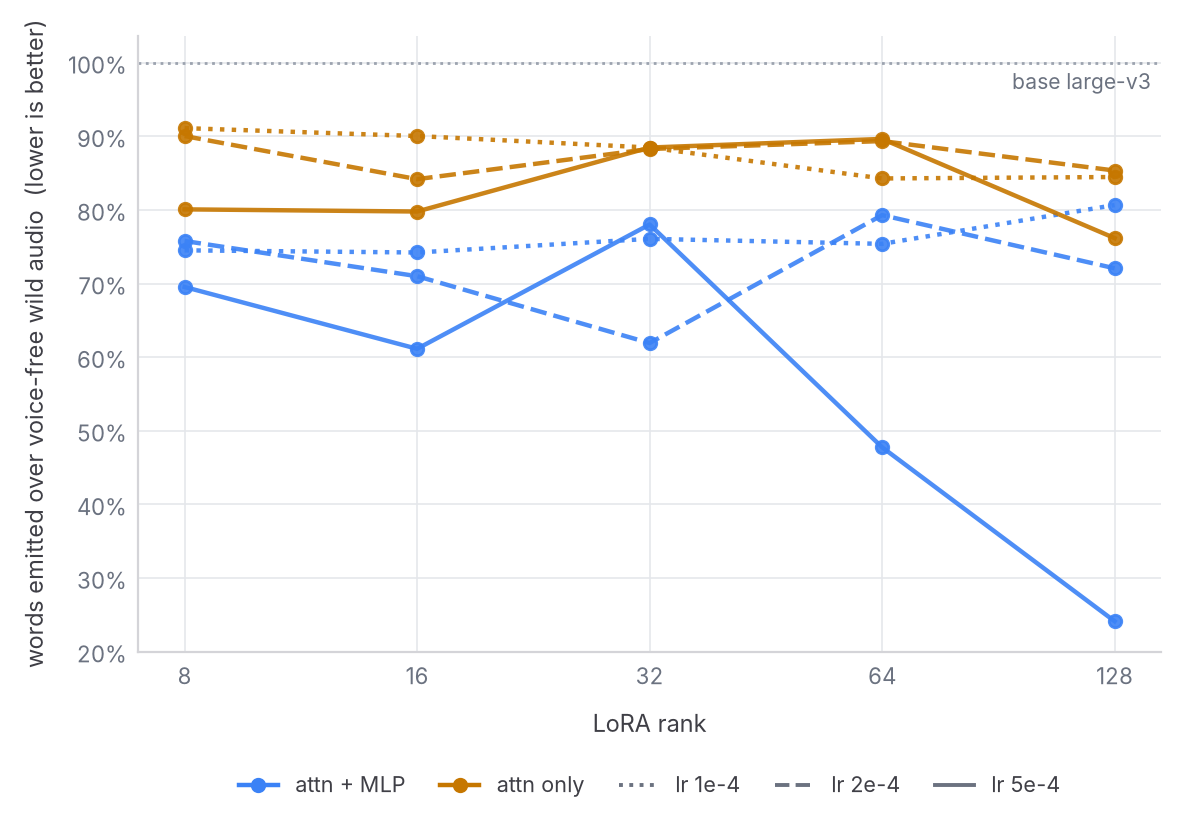}
  \caption{Same rank, same learning rate, same mix, different modules. Adapting the two
  feed-forward matrices as well as the four attention projections is what makes the model stop.
  Attention-only adapters barely move off the base model at any rank.}
  \label{fig:sweepfamily}
\end{figure}

Figure~\ref{fig:sweepfamily} isolates the module choice: rank on the horizontal axis, one
line per learning rate, the two families in different colours, everything else held fixed.

\textbf{Learning to stop lives in the feed-forward layers.} On the \texttt{all} mix, across all fifteen matched rank and learning-rate cells, adapting
\texttt{fc1} and \texttt{fc2} as well as the attention projections lowers word emission on
wild audio in \textbf{15 of 15}, by a median of 0.133 (Figure~\ref{fig:sweepfamily}). Attention-only adapters span 0.762 to 0.911 across the whole rank range, against a
base model at 0.999: five ranks and three learning rates, and none of them really learns to be
quiet. Adding the MLP pair spans 0.241 to 0.807 over the same grid.

Capacity is not the lever, coverage is. Attention plus MLP at rank 8, which is 14.4 M trainable
parameters, reaches 0.695 on wild audio. Attention only at rank 64, which is 62.9 M, reaches
0.896. Four and a half times the parameters in the wrong places is worse than a
quarter of them in the right ones.

\textbf{And the MLP only helps when the positives are there.} Run the same comparison on
\texttt{no\_synth} and it disappears: attention plus MLP wins 7 of 15 with a median difference
of $+0.005$, which is noise. The extra capacity is what lets the model use the contrastive
signal; without the positives there is nothing for it to use it on.

\textbf{One run diverged, and the positives are what prevented it.} Attention-only at rank 128
and 5e-4 on \texttt{no\_synth} finished at loss 8.27 and is destroyed: LibriSpeech word error
rate 0.985, phrase recovery 0.000, words on 100\% of silence. Its twin, the identical
configuration on \texttt{all}, converged to 0.658 and is a perfectly ordinary checkpoint at
0.762 on wild audio and 0.040 on LibriSpeech. Same rank, same learning rate, same steps. That
pair is one of the 33 and we have left it in, because excluding a run for diverging on one arm
of a paired comparison is exactly the filtering this paper argues against; with it removed the
headline is 30 of 32 rather than 31 of 33.

\textbf{Repetition on wild audio does not move.} The \texttt{wild loop} column sits between
0.000 and 0.017 for every run against a base of 0.008, and \texttt{all} is lower in only 6 of
33 pairs. Real voice-free clips rarely loop in the first place, so there is almost nothing
there to fix. The
repetition effect is real but it lives on the \texttt{reduplication} arm, where the deletion
rate never exceeds 0.006 across all 66 runs, meaning no configuration bought its quiet by
erasing repeated speech.

\textbf{Every fine-tune is worse than the base model on FLEURS.} The best of the 24 is 0.381
against the base model's 0.305. The synthetic positives cut the damage substantially, median
0.460 against 0.607, but they do not remove it. The corpus is 75 languages and FLEURS is 58, and
a thousand steps at a Malay-weighted mix moves the model away from the rest. This is the
honest cost of the method and it is the first thing we would attack next. Without this arm we
would not have known: LibriSpeech word error rate is flat at 0.034 to 0.037 across almost every
run in the table.

\section{Conclusion}

Whisper's habit of writing sentences over silence is in the weights, not in the decoding loop.
That matters because the usual remedies live outside the weights. A voice activity detector, a
no-speech threshold, a blocklist: all of them help, and all of them disappear the moment the
model is served on a stack that does not implement Whisper's own transcription loop.

It also matters because Whisper is copied. Distil-Whisper, Kotoba-Whisper and Granary are all
built on its pseudo-labels, and Canary-1B-v2 and Parakeet-TDT-0.6B-v3 are built on Granary. The
pipelines that produce them filter hallucinated transcripts out of the corpus, which is the
right thing to do for corpus quality and the wrong thing to do for this problem: it deletes the
evidence and keeps the behaviour. The student is fitted to the half of the teacher that was
already correct.

Adding non-speech audio so the model learns to stay quiet is already standard, and on its own
it does not work. The checkpoint in our benchmark that is best on every non-speech arm,
16.7\% on silence against 61.9\% for base \texttt{large-v3}, is also the one that recovers the
fewest genuinely spoken phrases, deletes 58\% of repeated speech, and returns character error
rate 1.26 on the typical FLEURS language while matching the base model on LibriSpeech. Our own
sweep reproduces that failure from scratch in a thousand steps whenever the negatives are
unaccompanied: the quietest run in the paper, 0.071 on real voice-free audio, recovers 29.2\%
of spoken phrases and triples LibriSpeech word error rate. Negative-only supervision teaches
suppression without teaching discrimination, and a model can always be made quiet by making it
mute.

The missing ingredient is cheap once you look for it. Hallucinations are a small, enumerable
set of phrases, so we collected 40,891 of them in 100 languages, chose a text-to-speech system
by measurement, and synthesised them as positives: the same text the negatives teach the model
to suppress, this time genuinely spoken and labelled with its transcript. In 33 matched pairs of fine-tunes that differ by nothing else, those clips lower word emission
on real voice-free audio in 31 pairs and raise phrase recovery in 32, and among the 30 pairs
that hold English accuracy it is 30 out of 30. The best checkpoint takes silence hallucination from
61.9\% to 2.4\% and words over real voice-free audio from 99.9\% to 47.8\% while raising phrase
recovery from 69.8\% to 82.7\%, for five thousandths of LibriSpeech word error rate. That is
the top-left corner Figure~\ref{fig:tradeoff} shows to be empty.

Three things we would not have known without measuring both halves on every run. Training loss
ranks these models backwards, because a blank-heavy mix is cheap to fit and the mix that trains
to the better loss is the worse model. Every fine-tune we produced is worse than the base model
on FLEURS while LibriSpeech stays flat at 0.034 to 0.037, so an English-only accuracy guard
would have reported no cost at all. And which modules the adapters touch matters more than how
many parameters they have: attention-only adapters do not learn to stop at any rank we tried,
while adding the two feed-forward matrices at rank 8, a quarter of the parameters, does. The benchmark, the lexicon, the synthetic corpus and
the wild audio are released so the next attempt can be scored the same way.

\section{Limitations}

The positive pool is largely synthetic. It is 52\% synthetic text-to-speech and 23\% read prose,
and only 25\% is real conversational speech, which is Sarawak dialect. A model tuned on
synthesised phrases may be learning something about the generator as well as about the phrases.

There is no telephony condition. Everything is 16 kHz, and a great deal of production audio is
8 kHz narrowband.

Chinese and Tamil positives are close to absent, 11 and 4 clips.

The \texttt{silence} arm is 42 clips, so its resolution is 2.4 percentage points per clip.
Differences of a few points on that arm are noise. The \texttt{music} arm may contain singing,
since its source labels vocals as unknown. \texttt{nonspeech} is the label-verified alternative
and should be preferred when the two disagree.

\texttt{speech\_in\_noise} is synthetic mixing, so the signal-to-noise ratio is exact but there
is no Lombard effect and no channel.

The synthetic corpus was filtered using \texttt{whisper-large-v3} with the language forced, so
that checkpoint is scored partly on clips it already agreed with. This flatters it against
\texttt{large-v2} and \texttt{turbo}. It does not explain a thirty point gap to the Malay
fine-tunes.

\section{Future work}

The decoder-side ablation is designed and not run: 34 configurations of beam width, temperature
fallback, repetition and presence penalties, and the interaction between those and a fine-tuned
checkpoint. The interesting question is whether a model trained to stop still needs the knobs.

Stage two of the sweep is the winning configuration, LoRA at rank 64 and 5e-4 on the
\texttt{all} mix, applied to \texttt{whisper-large-v3-turbo}. That is the checkpoint most
people actually serve and the one with the worst repetition profile of the three base models.
Every run here is a thousand steps, which is roughly a third of an epoch of the mix; whether the
FLEURS regression is a transient of undertraining or a real cost of the data is the first thing
a longer run would settle.

Two languages, Sinhala and Malagasy, have no FLEURS configuration, so they have no judge floor
and fall back to a flat gate that accepts almost nothing. They need a floor measured from
another corpus before their phrases can be synthesised reliably.

Finally, the wild pool should grow. AudioSet yields more than half of what it is asked for, so
collecting ten times as much costs only compute and storage. At that size hallucination could
be measured per acoustic condition instead of in aggregate.

\section{Acknowledgement}

This work was carried out at Scicom (MSC) Berhad. The author thanks Scicom for the compute and
for the time to publish the negative results alongside the positive ones.

\bibliographystyle{plainnat}
\bibliography{neurips_2023}

\begin{thebibliography}{26}
\providecommand{\natexlab}[1]{#1}
\providecommand{\url}[1]{\texttt{#1}}
\expandafter\ifx\csname urlstyle\endcsname\relax
  \providecommand{\doi}[1]{doi: #1}\else
  \providecommand{\doi}{doi: \begingroup \urlstyle{rm}\Url}\fi

\bibitem[Arbonel(2024)]{arbonel2024hfhalluc}
Sacha Arbonel.
\newblock whisper-hallucinations: a collected list of whisper hallucination
  phrases.
\newblock Hugging Face dataset, 2024.
\newblock URL
  \url{https://huggingface.co/datasets/sachaarbonel/whisper-hallucinations}.

\bibitem[Bara\'nski et~al.(2025)Bara\'nski, Jasi\'nski, Bartolewska, Kacprzak,
  Witkowski, and Kowalczyk]{baranski2025nonspeech}
Mateusz Bara\'nski, Jan Jasi\'nski, Julitta Bartolewska, Stanis{\l}aw Kacprzak,
  Marcin Witkowski, and Konrad Kowalczyk.
\newblock Investigation of whisper asr hallucinations induced by non-speech
  audio, 2025.
\newblock URL \url{https://arxiv.org/abs/2501.11378}.

\bibitem[Carletta et~al.(2005)Carletta, Ashby, Bourban, Flynn, Guillemot, Hain,
  Kadlec, Karaiskos, Kraaij, Kronenthal, Lathoud, Lincoln, Lisowska, McCowan,
  Post, Reidsma, and Wellner]{carletta2005ami}
Jean Carletta, Simone Ashby, Sebastien Bourban, Mike Flynn, Mael Guillemot,
  Thomas Hain, Jaroslav Kadlec, Vasilis Karaiskos, Wessel Kraaij, Melissa
  Kronenthal, Guillaume Lathoud, Mike Lincoln, Agnes Lisowska, Iain McCowan,
  Wilfried Post, Dennis Reidsma, and Pierre Wellner.
\newblock The ami meeting corpus: A pre-announcement.
\newblock In \emph{International Workshop on Machine Learning for Multimodal
  Interaction}, 2005.

\bibitem[Chen et~al.(2021{\natexlab{a}})Chen, Chai, Wang, Du, Zhang, Weng, Su,
  Povey, Trmal, Zhang, Jin, Khudanpur, Watanabe, Zhao, Zou, Li, Yao, Wang,
  Wang, You, and Yan]{chen2021gigaspeech}
Guoguo Chen, Shuzhou Chai, Guanbo Wang, Jiayu Du, Wei-Qiang Zhang, Chao Weng,
  Dan Su, Daniel Povey, Jan Trmal, Junbo Zhang, Mingjie Jin, Sanjeev Khudanpur,
  Shinji Watanabe, Shuaijiang Zhao, Wei Zou, Xiangang Li, Xuchen Yao, Yongqing
  Wang, Yujun Wang, Zhao You, and Zhiyong Yan.
\newblock Gigaspeech: An evolving, multi-domain asr corpus with 10,000 hours of
  transcribed audio, 2021{\natexlab{a}}.
\newblock URL \url{https://arxiv.org/abs/2106.06909}.

\bibitem[Chen et~al.(2021{\natexlab{b}})Chen, Wang, Chen, Wu, Liu, Chen, Li,
  Kanda, Yoshioka, Xiao, Wu, Zhou, Ren, Qian, Qian, Wu, Zeng, and
  Wei]{chen2021wavlm}
Sanyuan Chen, Chengyi Wang, Zhengyang Chen, Yu~Wu, Shujie Liu, Zhuo Chen, Jinyu
  Li, Naoyuki Kanda, Takuya Yoshioka, Xiong Xiao, Jian Wu, Long Zhou, Shuo Ren,
  Yanmin Qian, Yao Qian, Jian Wu, Michael Zeng, and Furu Wei.
\newblock Wavlm: Large-scale self-supervised pre-training for full stack speech
  processing, 2021{\natexlab{b}}.
\newblock URL \url{https://arxiv.org/abs/2110.13900}.

\bibitem[Conneau et~al.(2022)Conneau, Ma, Khanuja, Zhang, Axelrod, Dalmia,
  Riesa, Rivera, and Bapna]{conneau2022fleurs}
Alexis Conneau, Min Ma, Simran Khanuja, Yu~Zhang, Vera Axelrod, Siddharth
  Dalmia, Jason Riesa, Clara Rivera, and Ankur Bapna.
\newblock Fleurs: Few-shot learning evaluation of universal representations of
  speech, 2022.
\newblock URL \url{https://arxiv.org/abs/2205.12446}.

\bibitem[Defferrard et~al.(2017)Defferrard, Benzi, Vandergheynst, and
  Bresson]{defferrard2017fma}
Michael Defferrard, Kirell Benzi, Pierre Vandergheynst, and Xavier Bresson.
\newblock Fma: A dataset for music analysis, 2017.
\newblock URL \url{https://arxiv.org/abs/1612.01840}.

\bibitem[Fonseca et~al.(2020)Fonseca, Favory, Pons, Font, and
  Serra]{fonseca2020fsd50k}
Eduardo Fonseca, Xavier Favory, Jordi Pons, Frederic Font, and Xavier Serra.
\newblock Fsd50k: An open dataset of human-labeled sound events, 2020.
\newblock URL \url{https://arxiv.org/abs/2010.00475}.

\bibitem[Galvez et~al.(2021)Galvez, Diamos, Ciro, Ceron, Achorn, Gopi, Kanter,
  Lam, Mazumder, and Reddi]{galvez2021peoplesspeech}
Daniel Galvez, Greg Diamos, Juan Ciro, Juan~Felipe Ceron, Keith Achorn, Anjali
  Gopi, David Kanter, Maximilian Lam, Mark Mazumder, and Vijay~Janapa Reddi.
\newblock The people's speech: A large-scale diverse english speech recognition
  dataset for commercial usage, 2021.
\newblock URL \url{https://arxiv.org/abs/2111.09344}.

\bibitem[Gandhi et~al.(2023)Gandhi, von Platen, and
  Rush]{gandhi2023distilwhisper}
Sanchit Gandhi, Patrick von Platen, and Alexander~M. Rush.
\newblock Distil-whisper: Robust knowledge distillation via large-scale pseudo
  labelling, 2023.
\newblock URL \url{https://arxiv.org/abs/2311.00430}.

\bibitem[Gemmeke et~al.(2017)Gemmeke, Ellis, Freedman, Jansen, Lawrence, Moore,
  Plakal, and Ritter]{gemmeke2017audioset}
Jort~F. Gemmeke, Daniel P.~W. Ellis, Dylan Freedman, Aren Jansen, Wade
  Lawrence, R.~Channing Moore, Manoj Plakal, and Marvin Ritter.
\newblock Audio set: An ontology and human-labeled dataset for audio events.
\newblock In \emph{IEEE International Conference on Acoustics, Speech and
  Signal Processing (ICASSP)}, 2017.

\bibitem[He et~al.(2024)He, Shang, Wang, Li, Gu, Hua, Liu, Yang, Li, Shi, Wang,
  Chen, Zhang, and Wu]{he2024emilia}
Haorui He, Zengqiang Shang, Chaoren Wang, Xuyuan Li, Yicheng Gu, Hua Hua, Liwei
  Liu, Chen Yang, Jiaqi Li, Peiyang Shi, Yuancheng Wang, Kai Chen, Pengyuan
  Zhang, and Zhizheng Wu.
\newblock Emilia: An extensive, multilingual, and diverse speech dataset for
  large-scale speech generation, 2024.
\newblock URL \url{https://arxiv.org/abs/2407.05361}.

\bibitem[Hu et~al.(2021)Hu, Shen, Wallis, Allen-Zhu, Li, Wang, Wang, and
  Chen]{hu2021lora}
Edward~J. Hu, Yelong Shen, Phillip Wallis, Zeyuan Allen-Zhu, Yuanzhi Li, Shean
  Wang, Lu~Wang, and Weizhu Chen.
\newblock Lora: Low-rank adaptation of large language models, 2021.
\newblock URL \url{https://arxiv.org/abs/2106.09685}.

\bibitem[Koenecke et~al.(2024)Koenecke, Choi, Mei, Schellmann, and
  Sloane]{koenecke2024careless}
Allison Koenecke, Anna Seo~Gyeong Choi, Katelyn~X. Mei, Hilke Schellmann, and
  Mona Sloane.
\newblock Careless whisper: Speech-to-text hallucination harms.
\newblock In \emph{Proceedings of the 2024 ACM Conference on Fairness,
  Accountability, and Transparency (FAccT)}, 2024.
\newblock URL \url{https://arxiv.org/abs/2402.08021}.
\newblock arXiv:2402.08021.

\bibitem[Koluguri et~al.(2025)Koluguri, Sekoyan, Zelenfroynd, Meister, Ding,
  Kostandian, Huang, Karpov, Balam, Lavrukhin, Peng, Papi, Gaido, Brutti, and
  Ginsburg]{koluguri2025granary}
Nithin~Rao Koluguri, Monica Sekoyan, George Zelenfroynd, Sasha Meister,
  Shuoyang Ding, Sofia Kostandian, He~Huang, Nikolay Karpov, Jagadeesh Balam,
  Vitaly Lavrukhin, Yifan Peng, Sara Papi, Marco Gaido, Alessio Brutti, and
  Boris Ginsburg.
\newblock Granary: Speech recognition and translation dataset in 25 european
  languages.
\newblock In \emph{Interspeech}, 2025.
\newblock URL \url{https://arxiv.org/abs/2505.13404}.
\newblock arXiv:2505.13404.

\bibitem[{Kotoba Technologies} and Ushio(2024)]{kotobawhisper2024}
{Kotoba Technologies} and Asahi Ushio.
\newblock Kotoba-whisper: Distilled whisper for japanese automatic speech
  recognition.
\newblock Hugging Face model card, 2024.
\newblock URL \url{https://huggingface.co/kotoba-tech/kotoba-whisper-v2.0}.

\bibitem[{Malaysia-AI}(2025)]{malaysiaai2025pipeline}
{Malaysia-AI}.
\newblock malaysia-ai/dataset: speech-to-text whisper data preparation
  pipeline.
\newblock GitHub repository, 2025.
\newblock URL
  \url{https://github.com/malaysia-ai/dataset/tree/main/stt-whisper}.

\bibitem[{Mesolitica}(2025)]{mesolitica2025stage2}
{Mesolitica}.
\newblock Malaysian-stt-whisper-stage2: annealing set for the malaysian whisper
  fine-tunes.
\newblock Hugging Face dataset, 2025.
\newblock URL
  \url{https://huggingface.co/datasets/mesolitica/Malaysian-STT-Whisper-Stage2}.

\bibitem[{Mesolitica} and {Malaysia-AI}(2025)]{mesolitica2025malaysianwhisper}
{Mesolitica} and {Malaysia-AI}.
\newblock Malaysian whisper: Open speech recognition fine-tunes and training
  pipeline for malaysian languages.
\newblock Hugging Face model and dataset cards; pipeline at
  \url{https://github.com/malaysia-ai/dataset/tree/main/stt-whisper}, 2025.
\newblock URL
  \url{https://huggingface.co/mesolitica/Malaysian-whisper-large-v3-turbo-v3}.

\bibitem[{Omnilingual ASR team}(2025)]{keren2025omnilingual}
{Omnilingual ASR team}.
\newblock Omnilingual asr: Open-source multilingual speech recognition for
  1600+ languages, 2025.
\newblock URL \url{https://arxiv.org/abs/2511.09690}.

\bibitem[Panayotov et~al.(2015)Panayotov, Chen, Povey, and
  Khudanpur]{panayotov2015librispeech}
Vassil Panayotov, Guoguo Chen, Daniel Povey, and Sanjeev Khudanpur.
\newblock Librispeech: An asr corpus based on public domain audio books.
\newblock In \emph{IEEE International Conference on Acoustics, Speech and
  Signal Processing (ICASSP)}, 2015.

\bibitem[Radford et~al.(2022)Radford, Kim, Xu, Brockman, McLeavey, and
  Sutskever]{radford2022whisper}
Alec Radford, Jong~Wook Kim, Tao Xu, Greg Brockman, Christine McLeavey, and
  Ilya Sutskever.
\newblock Robust speech recognition via large-scale weak supervision, 2022.
\newblock URL \url{https://arxiv.org/abs/2212.04356}.

\bibitem[Rio et~al.(2022)Rio, Ha, McNamara, Miller, and
  Chandra]{delrio2022earnings22}
Miguel~Del Rio, Peter Ha, Quinten McNamara, Corey Miller, and Shipra Chandra.
\newblock Earnings-22: A practical benchmark for accents in the wild, 2022.
\newblock URL \url{https://arxiv.org/abs/2203.15591}.

\bibitem[{Seamless Communication team}(2023)]{seamless2023m4t}
{Seamless Communication team}.
\newblock Seamlessm4t: Massively multilingual \& multimodal machine
  translation, 2023.
\newblock URL \url{https://arxiv.org/abs/2308.11596}.

\bibitem[Sekoyan et~al.(2025)Sekoyan, Koluguri, Tadevosyan, Zelasko, Bartley,
  Karpov, Balam, and Ginsburg]{sekoyan2025canaryv2}
Monica Sekoyan, Nithin~Rao Koluguri, Nune Tadevosyan, Piotr Zelasko, Travis
  Bartley, Nick Karpov, Jagadeesh Balam, and Boris Ginsburg.
\newblock Canary-1b-v2 \& parakeet-tdt-0.6b-v3: Efficient and high-performance
  models for multilingual asr and ast, 2025.
\newblock URL \url{https://arxiv.org/abs/2509.14128}.

\bibitem[Wang et~al.(2021)Wang, Riviere, Lee, Wu, Talnikar, Haziza, Williamson,
  Pino, and Dupoux]{wang2021voxpopuli}
Changhan Wang, Morgane Riviere, Ann Lee, Anne Wu, Chaitanya Talnikar, Daniel
  Haziza, Mary Williamson, Juan Pino, and Emmanuel Dupoux.
\newblock Voxpopuli: A large-scale multilingual speech corpus for
  representation learning, semi-supervised learning and interpretation, 2021.
\newblock URL \url{https://arxiv.org/abs/2101.00390}.

\end{thebibliography}

\end{document}